\documentclass[10pt,twocolumn,letterpaper]{article}
\pdfoutput=1

\usepackage[breaklinks=true,bookmarks=false,letterpaper=true,colorlinks]{hyperref}
\usepackage{cvpr}
\usepackage{times}
\usepackage{graphicx}
\usepackage{amsmath,amssymb} 
\usepackage{color}

\usepackage{array}
\usepackage{cite}
\usepackage{multirow}
\usepackage{rotating}

\cvprfinalcopy

\graphicspath{{figures/}{figures/candidates/}{figures/cor_loc/}}
\begin{document}
\newcommand{\twofigures}[3]{
            \centerline{{\includegraphics[width=#3]{#1}}~~{\includegraphics[width=#3]{#2}}}}

\newcommand{\threefiguresh}[4]{
            \centerline{{\includegraphics[height=#4]{#1}}{\includegraphics[height=#4]{#2}}{\includegraphics[height=#4]{#3}}}}

\newcommand{\threefigures}[4]{
            \centerline{{\includegraphics[width=#4]{#1}}~~{\includegraphics[width=#4]{#2}}~~{\includegraphics[width=#4]{#3}}}}

\newcommand{\threefigureslbl}[5]{
            \centerline{(#5){\includegraphics[width=#4]{#1}}~~{\includegraphics[width=#4]{#2}}~~{\includegraphics[width=#4]{#3}}}}

\newcommand{\lthreefigures}[4]{
            \centerline{{\includegraphics[width=#4]{#1}}{\includegraphics[width=#4]{#2}}{\includegraphics[width=#4]{#3}}
			\makebox[0.33\columnwidth][c]{(a)}\makebox[0.33\columnwidth][c]{(b)}\makebox[0.33\columnwidth][c]{(c)}}}

\newcommand{\fourfigures}[5]{
            \centerline{{\includegraphics[width=#5]{#1}}~~{\includegraphics[width=#5]{#2}}~~{\includegraphics[width=#5]{#3}}~~{\includegraphics[width=#5]{#4}}}}

\newcommand{\sixfigures}[7]{
            \centerline{{\includegraphics[width=#7]{#1}}~~{\includegraphics[width=#7]{#2}}~~{\includegraphics[width=#7]{#3}}~~{\includegraphics[width=#7]{#4}}~~{\includegraphics[width=#7]{#5}}~~{\includegraphics[width=#7]{#6}}}}
            
\newcommand{\sixfiguresdots}[7]{
            \centerline{{\includegraphics[width=#7]{#1}}~~{\includegraphics[width=#7]{#2}}~~{\includegraphics[width=#7]{#3}}~~{...}~~{\includegraphics[width=#7]{#4}}~~{\includegraphics[width=#7]{#5}}~~{\includegraphics[width=#7]{#6}}}}

\newcommand{\threefigurescaption}[7]{
            \centerline{{\includegraphics[width=#4]{#1}}~~{\includegraphics[width=#4]{#2}}~~{\includegraphics[width=#4]{#3}}}
		     \makebox[#4][c]{#5}\makebox[#4][c]{#6}\makebox[#4][c]{#7}}

\newcommand{\comment}[1]{}

\newcommand{\twofigurescaption}[6]{
            \centerline{{\includegraphics[width=#3]{#1}}~~{\includegraphics[width=#3]{#2}}}
            \makebox[#6][c]{#4}\makebox[#6][c]{#5}}

\newcommand{\todo}[1]{{\bf TODO:} #1}

\newcolumntype{C}[1]{>{\centering}m{#1}}
\newcolumntype{P}[1]{p{#1\columnwidth}}
\def\etal{\emph{et~al}.~}
\newcommand{\F}[1]{Fig.~\ref{#1}}
\newcommand{\rottext}[1]{\begin{sideways}#1\end{sideways}}

\title{Weakly-Supervised Semantic Segmentation using Motion Cues}

\author{Pavel Tokmakov\thanks{Thoth team, Inria, Laboratoire Jean Kuntzmann, Grenoble, France.}
\and
Karteek Alahari\footnotemark[1]\vspace{0.3cm}\\
\large{Inria}
\and
Cordelia Schmid\footnotemark[1]\\
}
\maketitle

\begin{abstract}
Fully convolutional neural networks (FCNNs) trained on a large number of images
with strong pixel-level annotations have become the new state of the art for
the semantic segmentation task. While there have been recent attempts to learn
FCNNs from image-level weak annotations, they need additional constraints, such
as the size of an object, to obtain reasonable performance.  To address this
issue, we present motion-CNN (M-CNN), a novel FCNN framework which incorporates
motion cues and is learned from video-level weak annotations. Our learning
scheme to train the network uses motion segments as soft constraints, thereby
handling noisy motion information. When trained on weakly-annotated videos, our
method outperforms the state-of-the-art approach~\cite{papandreou2015weakly} on
the PASCAL VOC 2012 image segmentation benchmark. We also demonstrate that the
performance of M-CNN learned with 150 weak video annotations is on par with
state-of-the-art weakly-supervised methods trained with thousands of images.
Finally, M-CNN substantially outperforms recent approaches in a related task of
video co-localization on the YouTube-Objects dataset.
\end{abstract}

\section{Introduction}
The need for weakly-supervised learning for semantic segmentation has been
highlighted recently~\cite{Vezhnevets12,pinheiro2014weakly,Hartmann12}. It is
particularly important, as acquiring a training set by labeling images manually
at the pixel level is significantly more expensive than assigning class labels
at the image level.  Recent segmentation approaches have used weak annotations
in several forms: bounding boxes around objects~\cite{Monroy12,WuJ14}, image
labels denoting the presence of a
category~\cite{pinheiro2014weakly,Vezhnevets12} or a combination of the
two~\cite{papandreou2015weakly}. All these previous approaches only use
annotation in images, i.e., bounding boxes, image tags, as a weak form of
supervision. Naturally, additional cues would come in handy to address this
challenging problem. As noted in~\cite{brox2010object}, motion is one such cue
for semantic segmentation, which helps us identify the extent of objects and
their boundaries in the scene more accurately. To our knowledge, motion has not
yet been leveraged for weakly-supervised semantic segmentation. In this work,
we aim to fill this gap by learning an accurate segmentation model with the
help of motion cues extracted from weakly-annotated videos.

\begin{figure}
\begin{center}
\twofigurescaption{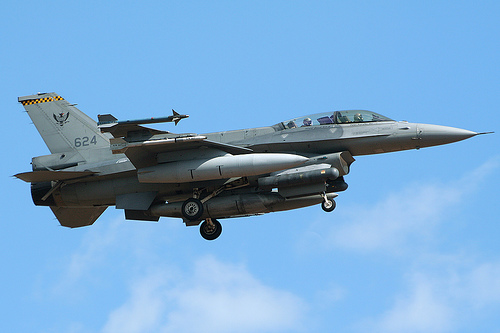}{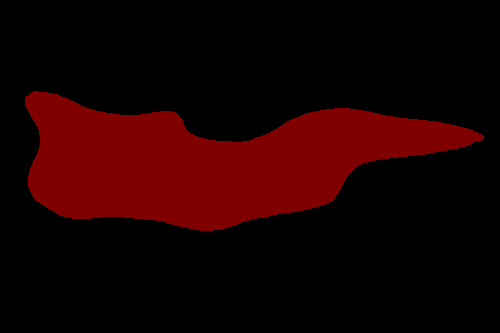}{0.45\columnwidth}{Image}{DeepLab~\cite{chen2014semantic} (full)}{0.5\columnwidth}
\twofigurescaption{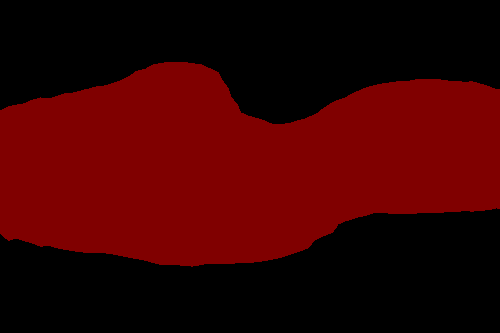}{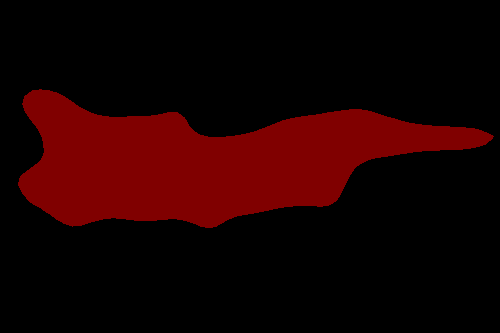}{0.45\columnwidth}{EM-Adapt~\cite{papandreou2015weakly} (weak)}{M-CNN (weak)}{0.5\columnwidth}
\end{center}
\caption{Comparison of state-of-the-art fully~\cite{chen2014semantic} and
weakly~\cite{papandreou2015weakly} supervised methods with our
weakly-supervised M-CNN model.}
\label{fig:teaser}
\end{figure}

Our proposed framework is based on fully convolutional neural networks
(FCNNs)~\cite{chen2014semantic,farabet2013learning,long2015fully,zheng2015conditional},
which extend deep CNNs, and are able to classify every pixel in an input image
in a single forward pass. While FCNNs show state-of-the-art results on
segmentation benchmark datasets, they require thousands of pixel-level
annotated images to train on---a requirement that limits their utility.
Recently, there have been some
attempts~\cite{pathak2014fully,pinheiro2014weakly,pathak2015constrained,papandreou2015weakly}
to train FCNNs with weakly-annotated images, but they remain inferior in
performance to their fully-supervised equivalents (see \F{fig:teaser}). In this
paper, we develop a new CNN variant named M-CNN, which leverages motion cues in
weakly-labeled videos, in the form of unsupervised motion segmentation,
e.g.,~\cite{papazoglou2013fast}. It builds on the architecture of FCNN by
adding a motion segmentation based label inference step, as shown in
\F{fig:overview}. In other words, predictions from the FCNN layers and motion
segmentation jointly determine the loss used to learn the network (see
\S\ref{sec:grabcut}).

Our approach uses unsupervised motion segmentation from real-world videos, such
as the YouTube-Objects~\cite{prest2012learning} and the
ImageNet-VID~\cite{imagenetvid} datasets, to train the network. In this
context, we are confronted with two main challenges. The first one is that even
the best-performing algorithms cannot produce good motion segmentations
consistently, and the second one is the ambiguity of video-level annotations,
which cannot guarantee the presence of object in all the frames. We develop a
novel scheme to address these challenges automatically without any manual
annotations, apart from the labels assigned at the video level, denoting the
presence of objects somewhere in the video. To this end, we use motion
segmentations as soft constraints in the learning process, and also fine-tune
our network with a small number of video shots to refine it.

We evaluated the proposed method on two related problems: semantic segmentation
and video co-localization. When trained on weakly-annotated videos, M-CNN
outperforms state-of-the-art EM-Adapt~\cite{papandreou2015weakly}
significantly, on the PASCAL VOC 2012 image segmentation
benchmark~\cite{pascalvoc2012}. Furthermore, our trained model, despite using
only 150 video labels, achieves performance similar to EM-Adapt trained on more
than 10,000 VOC image labels. Augmenting our training set with 1,000 VOC images
results in a further gain, achieving the best performance on VOC 2012 test set
in the weakly-supervised setting (see \S\ref{sec:exp_imvid}). On the video
co-localization task, where the goal is to localize common objects in a set of
videos, M-CNN substantially outperforms a recent method~\cite{Kwak15} by over
16\% on the YouTube-Objects dataset.
\begin{figure*}[t]
\centering
\includegraphics[width=1.8\columnwidth]{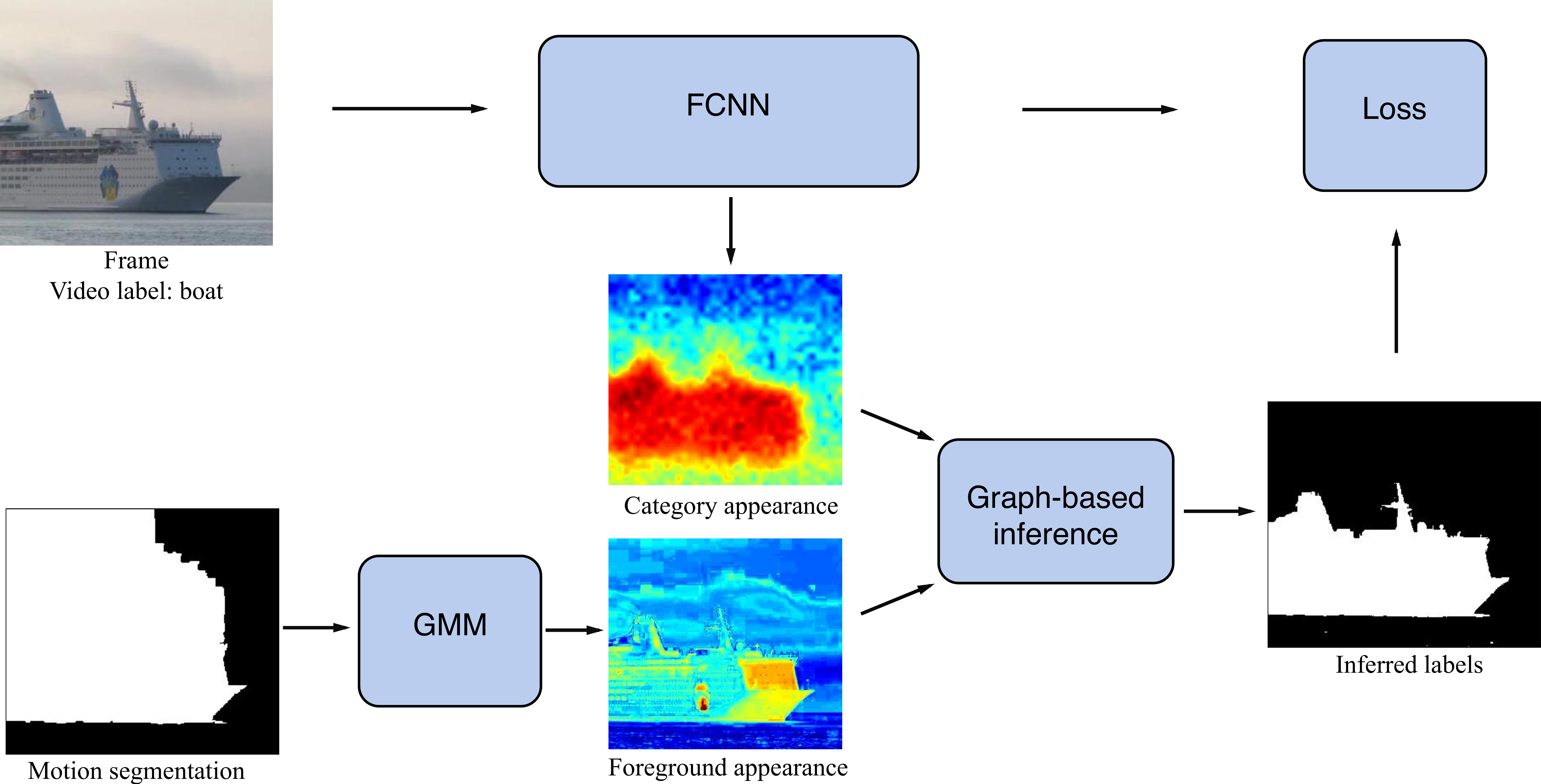}
\caption{Overview of our M-CNN framework, where we show only one frame from a
video example for clarity. The soft potentials (foreground appearance) computed
from motion segmentation and the FCNN predictions (category appearance) jointly
determine the latent segmentation (inferred labels) to compute the loss, and
thus the network update.}
\label{fig:overview}
\end{figure*}

The contributions of this work are twofold: (i)~We present a novel CNN
framework for segmentation that integrates motion cues in video as soft
constraints. (ii)~Experimental results show that our segmentation model learned
from weakly-annotated videos can indeed be applied to evaluate on challenging
benchmarks and achieves top performance on semantic segmentation as well as
video co-localization tasks. Code for training our M-CNN iteratively is
integrated in an FCNN framework in Caffe~\cite{jia2014caffe}, and will be made
available.

\section{Related Work}
In addition to fully-supervised segmentation approaches, such
as~\cite{carreira2012cpmc,carreira2012semantic}, several weakly-supervised
methods have been proposed over the years: some of them use bounding
boxes~\cite{Monroy12,WuJ14}, while others rely on image
labels~\cite{Vezhnevets12}. Traditional approaches for this task, such
as~\cite{Vezhnevets12}, used a variety of hand-crafted visual features, namely,
SIFT histograms, color, texture, in combination with a graphical or a
parametric structured model. Such early attempts have been recently
outperformed by FCNN methods, e.g.,~\cite{papandreou2015weakly}. 
 
FCNN architecture~\cite{chen2014semantic,farabet2013learning,long2015fully,zheng2015conditional,papandreou2015weakly,pathak2015constrained,pathak2014fully,pinheiro2014weakly,lin2015efficient} 
adapts standard CNNs~\cite{LeCun89,krizhevsky2012imagenet} to handle input
images of any arbitrary size by treating the fully connected layers as
convolutions with kernels of appropriate size. This allows them to output
scores for every pixel in the image. Most of these
methods~\cite{chen2014semantic,farabet2013learning,long2015fully,zheng2015conditional,lin2015efficient}
rely on strong pixel-level annotation to train the network.

Attempts~\cite{papandreou2015weakly,pathak2015constrained,pathak2014fully,pinheiro2014weakly}
to learn FCNNs for the weakly-supervised case use either a multiple instance
learning (MIL) scheme~\cite{pathak2014fully,pinheiro2014weakly} or constraints
on the distribution of pixel
labels~\cite{papandreou2015weakly,pathak2015constrained} to define the loss
function. For example, Pathak~\etal\cite{pathak2014fully} extend the MIL
framework used for object detection~\cite{cinbis2014multi,Russakovsky12} to
segmentation by treating the pixel with the highest prediction score for a
category as its positive sample when computing the loss. Naturally, this
approach is susceptible to standard issues suffered by MIL, like converging to
the most discriminative parts of objects~\cite{cinbis2014multi}. An alternative
MIL strategy is used in~\cite{pinheiro2014weakly}, by introducing a soft
aggregation function that translates pixel-level FCNN predictions into an image
label distribution. The loss is then computed with respect to the image
annotation label and backpropagated to update the network parameters. This
strategy works better in practice than~\cite{pathak2014fully}, but requires
training images that contain only a single object, as well as explicit
background images. Furthermore, it uses a complex post-processing step
involving multi-scale segmentations when testing, which is critical to its
performance.

Weakly-supervised FCNNs in~\cite{papandreou2015weakly,pathak2015constrained}
define constraints on the predicted pixel labels.
Papandreou~\etal\cite{papandreou2015weakly} presented an expectation
maximization (EM) approach, which alternates between predicting pixel labels
(E-step) and estimating FCNN parameters (M-step). Here, the label prediction
step is moderated with cardinality constraints, i.e., at least 20\% of the
pixels in an image need to be assigned to each of the image-label categories,
and at least 40\% to the background. This approach was extended
in~\cite{pathak2015constrained} to include generic linear constraints on the
label space, by formulating label prediction as a convex optimization problem.
Both these methods showed excellent results on the PASCAL VOC 2012 dataset, but
are sensitive to the linear/cardinality constraints. We address this drawback
in our M-CNN framework, where motion cues act as more precise constraints.
\F{fig:teaser} shows the improvement due to these constraints. We demonstrate
that FCNNs can be trained with videos, unlike all the previous methods
restricted to images, and achieve the best performance using much less training
data more effectively.

Weakly-supervised learning is also related to webly-supervised learning.
Methods following this recent
trend~\cite{chen2013neil,divvala2014learning,chen2015webly,liang2015computational}
are kick-started with either a small number of manually annotated examples,
e.g., some fully-supervised training examples for the object detection task
in~\cite{liang2015computational}, or automatically discovered ``easy''
samples~\cite{chen2015webly}, and then trained with a gradually increasing set
of examples mined from web resources. However, none of them address the
semantic segmentation problem. Other paradigms related to weakly-supervised
learning, such as co-localization~\cite{prest2012learning} and
co-segmentation~\cite{Rother06} require the video (or image) to contain a
dominant object class. Co-localization methods aim to localize the common
object with bounding boxes, whereas in co-segmentation, the goal is to estimate
pixel-wise segment labels. Such approaches,
e.g.,~\cite{Joulin14,prest2012learning,Tang13}, typically rely on a
pre-computed candidate set of regions (or boxes) and choose the best one with
an optimization scheme. Thus, they have no end-to-end learning mechanism and
are inherently limited by the quality of the candidates.

\section{Learning semantic segmentation from video}
\label{sec:method}
We begin by presenting a summary of the entire approach in
Section~\ref{sec:overview}. We then describe the network architecture in
Section~\ref{sec:net}, explain the estimation of latent segmentation variables
and the computation of the loss function for learning the network in
Section~\ref{sec:grabcut}, Finally, Section~\ref{sec:inclearn} presents the
fine-tuning step to further improve our M-CNN.

\subsection{Overview}
\label{sec:overview}
We train our network by exploiting motion cues from video sequences.
Specifically, we extract unsupervised motion segments from video, with
algorithms such as~\cite{papazoglou2013fast}, and use them in combination with
the weak labels at the video level to learn the network. We sample frames from
all the video sequences uniformly, and assign them the class label of the
video. This collection forms our training dataset, along with their
corresponding motion segments.

The parameters of M-CNN are updated with a standard mini-batch SGD, similar to
other CNN approaches~\cite{papandreou2015weakly}, with the gradient of a loss
function. Here, the loss measures the discrepancy between the ground truth
segmentation label and the label predicted at each pixel. Thus, in order to
learn the network for the semantic segmentation task, we need pixel-level
ground truth for all the training data. These pixel-level labels are naturally
latent variables in the context of weakly-supervised learning. Now, the task is
to estimate them for our weakly-labeled videos. An ideal scenario in this
setting would be near-perfect motion segmentations, which can be directly used
as object ground truth labels. However, in practice, not only are the
segmentations far from perfect (see \F{fig:mseg_good_bad}), but also fail to
capture moving objects in many of the shots. This makes a direct usage of
motion segmentation results suboptimal. To address this, we propose a novel
scheme, where motion segments are only used as soft constraints to estimate the
latent variables together with object appearance cues.

The other challenges when dealing with real-world video datasets, such as
YouTube-Objects and ImageNet-VID, are related to the nature of video data
itself. On one hand, not all parts of a video contain the object of interest.
For instance, a video from a show reviewing boats may contain shots with the
host talking about the boat, and showing it from the inside for a significant
part---content that is unsuitable for learning a segmentation model for the VOC
`boat' category. On the other hand, a long video can contain many nearly
identical object examples which leads to an imbalance in the training set. We
address both problems by fine-tuning our M-CNN with an automatically
selected, small subset of the training data.

\subsection{Network architecture}
\label{sec:net}
Our network is built on the DeepLab model for semantic image
segmentation~\cite{chen2014semantic}. It is an FCNN, obtained by converting the
fully-connected layers of the VGG-16 network~\cite{simonyan2014very} into
convolutional layers. A few other changes are implemented to get a dense
network output for an image at its full resolution efficiently. Our work builds
on this network. We develop a more principled and effective label prediction
scheme involving motion cues to estimate the latent variables, in contrast to
the heuristic size constraints used in~\cite{papandreou2015weakly}, which is
based on DeepLab.

\begin{figure}[t]
\centering
\threefigures{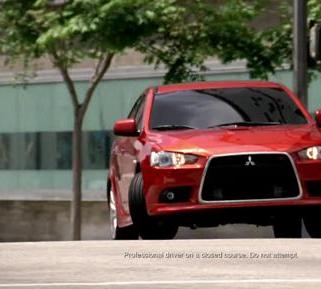}{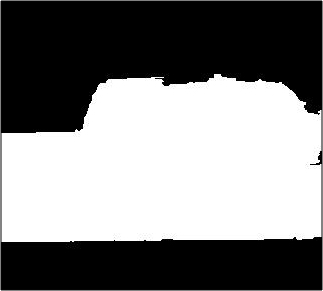}{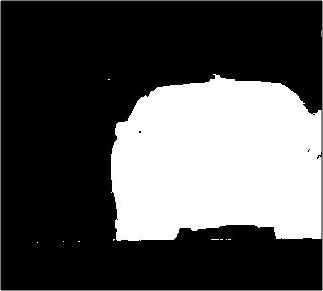}{0.3\columnwidth}
\vspace{0.05cm}
\threefigures{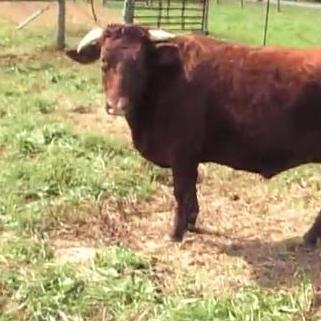}{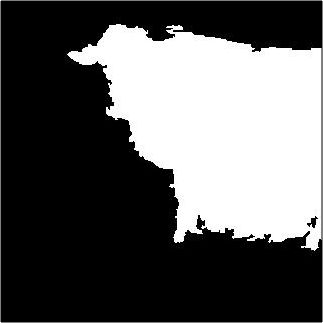}{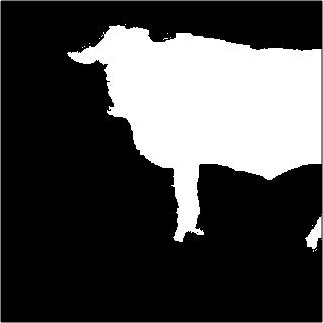}{0.3\columnwidth}
\vspace{0.05cm}
\threefigures{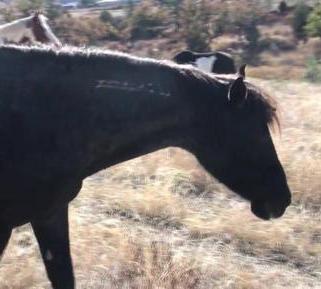}{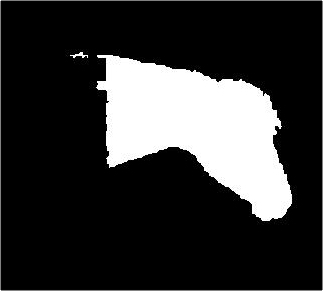}{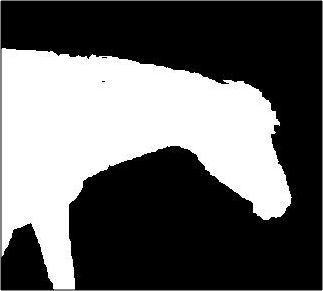}{0.3\columnwidth}
\makebox[0.3\columnwidth][c]{Frame~~~}\makebox[0.3\columnwidth][c]{Motion seg.~\cite{papazoglou2013fast}}\makebox[0.3\columnwidth][c]{~~~~~Label prediction}
\caption{Examples highlighting the importance of label prediction for handling
imprecise motion segmentations (second column). The soft GMM potentials
computed from motion segments together with network predictions produce better
labels (third column) to learn the network. See \S\ref{sec:grabcut} for
details.}
\label{fig:mseg_good_bad}
\end{figure}

\subsection{Estimating latent variables with label prediction}
\label{sec:grabcut}
Given an image of $N$ pixels, let 
${\bf p}$ denote the output of the softmax layer of the convolutional
network. Then, $p_i^l \in [0,1]$ is the prediction score of the network at
pixel $i$ for label $l$. The parameters of the network are updated
with the gradient of the loss function, given by:
\begin{equation}
\mathcal{L}({\bf x}, {\bf p}) = \sum_{i=1}^{N} \sum_{l=0}^{L} \delta(x_i - l) \log(p_i^l),
\label{eqn:origloss}
\end{equation}
where ${\bf x}$ denotes ground truth segmentation labels in the
fully-supervised case, ${\bf p}$ is the current network prediction, and
$\delta(x_i - l)$ is the Dirac delta function, i.e., $\delta(x_i - l) = 1$, if
$x_i = l$, and $0$ otherwise. The segmentation label $x_i$ of pixel $i$ takes
values from the label set ${\bf L} = \{0, 1,\ldots, L\}$, containing the
background class (0) and $L$ object categories. Naturally, in the
weakly-supervised case, ground truth segmentation labels are unavailable, and
${\bf x}$ represents latent segmentation variables, which need to be estimated.
We perform this estimation with soft motion segmentation cues in this paper.

Given the motion segmentation \mbox{${\bf s} = \{s_i
| i = 1,\ldots,N\}$,} where $s_i \in \{0,1\}$ denotes whether a pixel $i$
belongs to foreground ($1$) or background ($0$).\footnote{We do not include an
index denoting the frame number in the video for brevity.} The regions assigned
to foreground can represent multiple object categories when the video is tagged with more
than one category label. A simple way of transforming motion segmentation labels
$s_i$ into latent semantic segmentation labels $x_i$ is with a hard
assignment, i.e., $x_i = s_i$. This hard assignment is limited to videos
containing a single category label, and also makes the assumption that motion
segments are accurate and can be used as they are. We will see in our
experiments that this performs poorly when using real-world video datasets
(cf.\ `M-CNN* hard' in Table~\ref{tbl:fov}). We address this by using motion
cues as soft constraints for estimating the label assignment ${\bf x}$ in the
following.

\paragraph{Inference of the segmentation ${\bf x}$.} We compute the pixel-level
segmentation ${\bf x}$ as the minimum of an energy function $E({\bf x})$ defined
by:
\begin{equation}
E({\bf x}) = \sum_{i \in \mathcal{V}} \left(\psi_i^m(z_i) + \alpha \psi_i^{fc}(p_i^{x_i})\right)~+~\sum_{(i,j) \in \mathcal{E}} \psi_{ij}(x_i,x_j),
\label{eqn:energy}
\end{equation} 
where $\mathcal{V} = \{1, 2, \ldots, N\}$ is the set of all the pixels, $z_i$
denotes the RGB color at pixel $i$ and the set $\mathcal{E}$ denotes all pairs
of neighboring pixels in the image. Unary terms $\psi_i^m$ and $\psi_i^{fc}$
are computed from motion cues and current predictions of the network
respectively, with $\alpha$ being a scalar parameter balancing their impact.
The pairwise term $\psi_{ij}$ imposes a smoothness over the label space.

The first unary term $\psi_i^m$ captures the appearance of all foreground
objects obtained from motion segments. To this end, we learn two Gaussian
mixture models (GMMs), one each for foreground and background, with RGB values
of pixel colors, similar to standard segmentation
methods~\cite{rother2004grabcut,papazoglou2013fast}. The foreground GMM is
learned with RGB values of all the pixels assigned to foreground in the motion
segmentation. The background GMM is learned in a similar fashion with the
corresponding background pixels. Given the RGB values of a pixel $i$,
$\psi_i^m(z_i)$ is given by the negative log-likelihood of the corresponding
GMM (background one for $l=0$ and foreground otherwise). Using motion cues to
generate this soft potential $\psi_i^m$ helps us alleviate the issue of
imperfect motion segmentation. The second unary term $\psi_i^{fc}$ represents
the learned category appearance model determined by the current network
prediction $p_i^{x_i}$ for pixel $i$, i.e., \mbox{$\psi_i^{fc}(p_i^{x_i}) =
-\log(p_i^{x_i})$}.

The pairwise term is based on a contrast-sensitive Potts
model~\cite{rother2004grabcut,Boykov01} as:
\small
\begin{equation}
\psi_{ij}(x_i,x_j) = \lambda (1 - \Delta(i,j)) (1 - \delta(x_i - x_j)) \frac{\exp(-\gamma ||z_i - z_j||^2)}{\text{dist}(i,j)},
\label{eqn:pairwise}
\end{equation}
\normalsize

where $z_i$ and $z_j$ are colors of pixels $i$ and $j$, $\lambda$ is a scalar
parameter to balance the order of magnitude of the pairwise term with respect
to the unary term, and $\gamma$ is a scalar parameter set to 0.5 as
in~\cite{papazoglou2013fast}. The function $\text{dist}(i,j)$ is the Euclidean
distance between pixels. The Dirac delta function \mbox{$\delta(x_i - x_j)$}
ensures that the pairwise cost is only applicable when two neighboring pixels
take different labels. In addition to this, we introduce the term $(1 -
\Delta(i,j))$, where $\Delta(i,j) = 1$ if pixels $i$ and $j$ both fall in the
boundary region around the motion segment, and $0$ otherwise. This accounts for
the fact that motion segments may not always respect color boundaries, and
allows the minimization algorithm to assign different labels to neighboring
pixels around motion edges.

We minimize the energy function (\ref{eqn:energy}) with an iterative
GrabCut-like~\cite{rother2004grabcut} approach, wherein we first apply the
alpha expansion algorithm~\cite{Boykov01alpha} to get a multi-label solution,
use it to re-estimate the (background and foreground) GMMs, and then repeat the
two steps for a few iterations. We highlight the importance of our label
prediction technique with soft motion-cue constraints in \F{fig:mseg_good_bad}.
Here, the original, binary motion predictions are imprecise (bottom two rows)
or incorrect (top row) in all the examples, whereas using them as soft
constraints in combination with the network prediction results in a more accurate
estimation of the latent segmentation variables.

\subsection{Fine-tuning M-CNN}
\label{sec:inclearn}
We learn an initial M-CNN model from all the videos in the dataset which have
sufficient motion information (see \S\ref{sec:implement} for implementation
details). To refine this model we add a fine-tuning step, which updates the
parameters of the network with a small set of unique and reliable video
examples. This set is built automatically by selecting one shot from each video
sequence, whose motion segment has the highest overlap (intersection over
union) score with the current M-CNN prediction. The intuition behind this
selection criterion is that our MCNN has already learned to discriminate
categories of interest from the background, and thus, its predictions will have
the highest overlap with precise motion segmentations.  This model refinement
leverages the most reliable exemplars and avoids near duplicates, often
occurring within one video. In Section~\ref{sec:exp_video} we demonstrate that
this step is necessary for dealing with real-world non-curated video data.

\section{Results and Evaluation}
\label{sec:expts}
\subsection{Experimental protocol}
We trained our M-CNN in two settings. The first one is on purely video data,
and the second on a combination of image and video data. We performed
experiments primarily with the weakly-annotated videos in the YouTube-Objects
v2.2 dataset~\cite{youtubeobjv2}. Additionally, to demonstrate that our
approach adapts to other datasets automatically, we used the ImageNet video
(ImageNet-VID) dataset~\cite{imagenetvid}. The weakly-annotated images to train
our network jointly on image and video data were taken from the training part
of the PASCAL VOC 2012 segmentation dataset~\cite{pascalvoc2012} with their
image tags only. We then evaluated variants of our method on the VOC 2012
segmentation validation and test sets.

\begin{table*}[t]
\begin{center}
\begin{tabular}{l|C{8mm}|C{6mm}C{6mm}C{6mm}C{6mm}C{6mm}C{6mm}C{6mm}C{6mm}C{6mm}C{6mm}C{6mm}|m{16mm}}
\hline
Method & FOV & bkg & aero & bird & boat & car & cat & cow & dog & horse & mbike & train & Average \\
\hline
EM-Adapt   & small & 65.7 & 25.1 & 20.5 & ~9.3 & 21.6 & 23.7 & 12.4 & 17.7 & 14.9 & 19.5 & 25.4 & 23.2 $\pm$ 3.0 \\
EM-Adapt   & large & 69.1 & 12.9 & 14.7 & ~9.0 & 12.9 & 15.4 & ~5.6 & ~9.9 & ~7.8 & 15.9 & 23.0 & 17.9 $\pm$ 4.4 \\
\hline
M-CNN*     & small & 83.4 & 30.3 & 35.2 & 13.5 & 11.6 & 36.5 & 22.1 & 19.8 & 22.2 & ~5.2 & 13.7 & 26.7 $\pm$ 1.0 \\
M-CNN*     & large & 84.6 & 35.3 & 44.8 & 24.7 & 21.7 & 44.4 & 26.3 & 26.5 & 27.9 & 10.0 & 22.9 & 33.6 $\pm$ 0.2 \\
M-CNN* hard& large & 83.6 & 35.3 & 38.6 & 24.0 & 21.2 & 39.6 & 20.2 & 21.3 & 19.2 & ~7.9 & 17.9 & 29.9 $\pm$ 0.7 \\
\hline
M-CNN      & large & 86.3 & 46.5 & 43.5 & 27.6 & 34.0 & 47.5 & 28.7 & 31.0 & 30.8 & 32.4 & 43.4 & 41.2 $\pm$ 1.3 \\
\hline
\end{tabular}
\vspace{0.2cm}
\caption{Performance of M-CNN and EM-Adapt variants, trained with
YouTube-Objects, on the VOC 2012 validation set. `*' denotes the M-CNN models
without fine-tuning. `M-CNN* hard' is the variant without the label prediction
step. `M-CNN' is our complete method: with fine-tuning and label prediction. We
report average and standard deviation over 5 runs.}
\label{tbl:fov}
\end{center}
\end{table*}

The YouTube-Objects dataset consists of 10 classes, with 155 videos in total.
Each video is annotated with one class label and is split automatically into
shots, resulting in 2511 shots overall. For evaluation, one frame per shot is
annotated with a bounding box in some of the shots. We use this exclusively for
evaluating our video co-localization performance in Section~\ref{sec:corloc}.
For experiments with ImageNet-VID, we use 795 training videos corresponding to
the 10 classes in common with YouTube-Objects. ImageNet-VID has bounding box
annotations produced semi-automatically for every frame in a video shot (2120
shots in total). We accumulate the labels over a shot and assign them as class
labels for the entire shot. As in the case of YouTube-Objects, we only use
class labels at the video level and none of the available additional
annotations.

The PASCAL VOC 2012 dataset has 20 foreground object classes and a background
category. It is split into 1464 training, 1449 validation and 1456 test images.
For experiments dealing with the subset of 10 classes in common with
YouTube-Objects (see the list in Table~\ref{tbl:fov}), we treat the remaining
10 from VOC as irrelevant classes. In other words, we exclude all the
training/validation images which contain only the irrelevant categories. This
results in 914 training and 909 validation images. In images that contain an
irrelevant class together with any of the 10 classes in YouTube-Objects, we
treat their corresponding pixels as background for evaluation. Some of the
state-of-art methods~\cite{papandreou2015weakly,pathak2015constrained} use an
augmented version of the VOC 2012 dataset, with over 10,000 additional training
images~\cite{HariharanICCV2011}. Naturally the variants trained on this large
dataset perform significantly better than those using the original VOC dataset.
We do not use this augmented dataset in our work, but report state-of-the-art
results due to our motion cues.

The segmentation performance of all the methods is measured as the intersection
over union (IoU) score of the predicted segmentation and the ground truth. We
compute IoU for each class as well as the average over all the classes,
including background, following standard
protocols~\cite{pascalvoc2012,papandreou2015weakly}. We also evaluate our
segmentation results in the co-localization setting with the CorLoc
measure~\cite{Joulin14,papazoglou2013fast,prest2012learning}, which is defined
as the percentage of images with IoU score, between ground truth and predicted
bounding boxes, more than 0.5.

\subsection{Implementation details}
\label{sec:implement}
\paragraph{Motion segmentation.}
In all our experiments we used~\cite{papazoglou2013fast}, a state-of-the-art
method for motion segmentation. We perform two pruning steps before training
the network. First, we discard all shots with less than 20 frames ($2 \times$
the batch size of our SGD training).  Second, we remove shots without relevant
motion information: (i) when there are nearly no motion segments, or (ii) a
significant part of the frame is assigned to foreground. We prune them out by a
simple criterion based on the size of the foreground segments. We keep only the
shots where the estimated foreground occupies between 2.5\% and 50\% of the
frame area in each frame, for at least 20 contiguous frames in the shot. In
cases where motion segmentation fails in the middle of a shot, but recovers
later, producing several valid sequences, we keep the longest one. These two
steps combined remove about a third of the shots, with 1675 and 1691 shots
remaining in YouTube-Objects and ImageNet-VID respectively. We sample 10 frames
uniformly from each of these remaining shots to train the network.

\begin{table*}[t]
\begin{center}
\begin{tabular}{l|l|C{6mm}C{6mm}C{6mm}C{6mm}C{6mm}C{6mm}C{6mm}C{6mm}C{6mm}C{8mm}C{6mm}|c}
\hline
Method & Dataset & bkg & aero & bird & boat & car & cat & cow & dog & horse & mbike & train & Average \\
\hline
EM-Adapt & YTube    & 65.7 & 25.1 & 20.5 & ~9.3 & 21.6 & 23.7 & 12.4 & 17.7 & 14.9 & 19.5 & 25.4 & 23.2$\dagger$ \\
EM-Adapt & ImNet    & 66.1 & 22.8 & 18.7 & 16.9 & 26.7 & 35.7 & 22.4 & 23.6 & 21.4 & 28.4 & 24.3 & 27.9~ \\
EM-Adapt & VOC      & 75.5 & 30.5 & 27.4 & 24.1 & 41.8 & 36.8 & 25.5 & 33.3 & 29.3 & 40.0 & 29.7 & 35.8~ \\
EM-Adapt & VOC aug. & 77.4 & 32.1 & 30.8 & 26.4 & 42.6 & 40.7 & 32.8 & 37.8 & 35.1 & 45.2 & 41.1 & 40.2~ \\
\hline
M-CNN    & YTube    & 86.3 & 46.5 & 43.5 & 27.6 & 34.0 & 47.5 & 28.7 & 31.0 & 30.8 & 32.4 & 43.4 & 41.2$\dagger$ \\
M-CNN    & VOC+YTube& 85.4 & 54.5 & 40.8 & 35.5 & 41.2 & 47.5 & 38.3 & 42.0 & 41.5 & 45.0 & 47.8 & 47.2$\dagger$ \\
M-CNN    & VOC aug.+YTube& 82.5 & 47.8 & 35.3 & 29.6 & 45.6 & 54.6 & 40.3 & 46.6 & 44.8 & 52.2 & 56.6 & 48.7~ \\
M-CNN    & ImNet    & 85.6 & 41.4 & 45.3 & 23.2 & 38.6 & 42.3 & 36.0 & 35.1 & 21.1 & 15.3 & 44.8 & 39.0~ \\
M-CNN    & VOC+ImNet& 85.1 & 53.3 & 46.8 & 32.5 & 33.9 & 37.3 & 40.7 & 32.3 & 34.2 & 40.0 & 45.0 & 43.7~ \\
M-CNN    & VOC aug.+ImNet& 83.1 & 47.6 & 40.3 & 26.4 & 44.1 & 51.1 & 41.7 & 51.0 & 34.9 & 44.6 & 52.7 & 47.0~ \\
\hline
\end{tabular}
\vspace{0.2cm}
\caption{Performance of our M-CNN variants on the VOC 2012 validation set is
shown as IoU scores. We also compare with the best variants of
EM-Adapt~\cite{papandreou2015weakly} trained on YouTube-Objects (YTube),
ImageNet-VID (ImNet), VOC, and augmented VOC (VOC aug.) datasets. $\dagger$
denotes the average result of 5 trained models.}
\label{tbl:results}
\end{center}
\end{table*}
\paragraph{Training.} 
We use a mini-batch of size 10 for SGD, where each mini-batch consists of the
10 frame samples of one shot. Our CNN learning parameters follow the
setting in~\cite{papandreou2015weakly}. The initial learning rate is set to 0.001 and
multiplied by 0.1 after a fixed number of iterations. We use a momentum of 0.9
and a weight decay of 0.0005. Also, the loss term $\delta(x_i - l) \log(p_i^l)$
computed for each object class $l$ with $\text{num}_l$ training samples, in
(\ref{eqn:origloss}), is weighted by $\min_{j=1 \dots
L}\text{num}_j/\text{num}_l$. This accounts for imbalanced number of training
samples for each class in the dataset.

In the energy function (\ref{eqn:energy}), the parameter $\alpha$, which
controls the relative importance of the current network prediction and the soft
motion cues, is set to 1 when training on the entire dataset. It is increased
to 2 for fine-tuning, where the predictions are more reliable due to an
improved network. We perform 4 iterations of the graph cut based inference
algorithm, updating the GMMs at each step. The inference algorithm is either
alpha expansion (for videos with multiple objects) or graph cut (when there is
only one object label for the video). Following~\cite{papazoglou2013fast}, we
learn GMMs for a frame $t$ with the motion segments from all the 10 frames in a
batch, weighting each of them inversely according to their distance from $t$.
The fine-tuning step is performed very selectively with the best shot for each
video, where the average overlap is no less than 0.2.

A systematic evaluation on the VOC 2012 validation set confirmed that the
performance is not sensitive to the number of iterations and the $\alpha$
parameter. The number of iterations is set as in other iterative graph cut
based methods, e.g.,~\cite{papazoglou2013fast}. In experiments on the VOC 2012
validation set, with the model trained on YouTube-Objects (M-CNN* in
Table~\ref{tbl:fov}), we found that this has a marginal impact on the
performance: changing the number of iterations from 1 through 5 resulted in
average IoU scores 33.6, 33.1, 33.5, 33.6 and 33.9 respectively. The $\alpha$
parameter is set based on the intuition that the network predictions are more
reliable in the fine-tuning step, where the network is already trained on the
entire dataset. The performance is again not sensitive within a range of
values, with only extreme cases changing IoU significantly: $\alpha$ = 0.5:
24.7, 1.0: 33.8, 2.0: 34.1, 3.0: 34.3, 10.0: 23.3. In the fine-tuning step
(M-CNN in Table~\ref{tbl:fov}), there is even less of an impact due to a better
trained model: $\alpha$ = 0.5: 41.4, 1.0: 41.9, 2.0: 42.3, 3.0: 42.6, 10.0:
42.2.

\paragraph{Code.} We implemented our M-CNN in the Caffe
framework~\cite{jia2014caffe}, with the proposed label prediction step as a new
layer. We will make our source code, configuration files, and trained models
available online~\cite{project}, to allow the reproduction of all the reported
results.

\subsection{Evaluation of M-CNN}
\label{sec:exp_video}
We start by evaluating the different components of our M-CNN approach and
compare to the state-of-the-art EM-Adapt method, see Table~\ref{tbl:fov}. We
train EM-Adapt and M-CNN with the pruned shots from our YouTube-Objects
training set in two network settings: large and small field of view (FOV). The
large FOV is 224$\times$224, while the small FOV is 128$\times$128. We learn 5
models which vary in the order of the training samples and their variations
(cropping, mirroring), and report the mean score and standard deviation.

The small FOV M-CNN without the fine-tuning step achieves an IoU of 26.7\%,
whereas large FOV gives 33.6\% on the PASCAL VOC 2012 validation set. In
contrast, EM-Adapt~\cite{papandreou2015weakly} trained\footnote{We used the
original implementation provided by the authors to train EM-Adapt.} on the same
dataset performs poorly with large FOV. Furthermore, both the variants of
EM-Adapt are lower in performance than our M-CNN, notably about 16\% for large
FOV. This is because EM-Adapt uses a heuristic (where background is constrained
to 40\% of the image area, and foreground to at least 20\%) to estimate the
latent segmentation labels, and fails to leverage the weak supervision in our
training dataset effectively. Our observation on this failure of EM-Adapt is
further supported by the analysis in~\cite{papandreou2015weakly}, which notes
that a large FOV network performs poorer than its small FOV counterpart when
only a ``small amount of supervision is leveraged''. The label prediction step
(\S\ref{sec:grabcut}) proposed in our method leverages training data better
than EM-Adapt, by optimizing an energy function involving soft motion
constraints and network responses. We also evaluated the significance of using
motion cues as soft constraints (M-CNN*) instead of introducing them as hard
labels (M-CNN* hard), i.e., directly using motion segmentation result as latent
labels ${\bf x}$. `M-CNN* hard' achieves 29.9 compared to 33.6 with soft
constraints. We then take our best variant (M-CNN with large FOV) and fine-tune
it, improving the performance further to 41.2\%. In all the remaining
experiments, we use the best variants of EM-Adapt and M-CNN.

\subsection{Training on weakly-annotated videos \& images}
\label{sec:exp_imvid}
We also trained our M-CNN with weakly-annotated videos and images. To this end,
we used images from the VOC 2012 training set. We added the 914 images from the
VOC 2012 training set containing the 10 classes, and used only their weak
annotations, i.e., image-level labels. In this setting, we first trained the
network with the pruned video shots from YouTube-Objects, fine-tuned it with a
subset of shots (as described in \S\ref{sec:inclearn}), and then performed a
second fine-tuning step with these selected video shots and VOC images. To
estimate the latent segmentation labels we use our optimization framework
(\S\ref{sec:grabcut}) when the training sample is from the video dataset and
the EM-Adapt label prediction step when it is from the VOC set. We can
alternatively use our framework with only the network prediction component for
images. For example, fine-tuning M-CNN with VOC and YouTube-Objects using the
network prediction component only for VOC images (i.e., without EM-Adapt)
improves the performance to 51.0 (from 47.2 in Table~\ref{tbl:results}). The
sucess of this depends on the quality of the network prediction, and it is not
viable when training on classes without video data, i.e., the remaining 10
classes in VOC.

\setlength{\tabcolsep}{4pt}
\begin{table*}[t]
\begin{center}
\begin{tabular}{l|lrcp{1.3cm}}
\hline
Method & Training data & \# samples & Average & Average 10-class \\
\hline
\multicolumn{4}{l}{\it Strong/Full supervision} & \\
\cite{pinheiro2014weakly} + bb         & VOC+ImNet& $\sim$762,500& 37.0 & 43.8 \\
\cite{pinheiro2014weakly} + seg        & VOC+ImNet& $\sim$761,500& 40.6 & 48.0 \\
\cite{papandreou2015weakly} + seg      & VOC aug. &       12,031 & 69.0 & 78.2 \\
\cite{mostajabi15} (full)              & VOC aug. &       10,582 & 69.6 & 79.3 \\
\cite{zheng2015conditional} (full)     & VOC aug.+COCO&   77,784 & 74.7 & 82.9 \\
\hline
\multicolumn{4}{l}{\it Weak supervision with additional info.} & \\
\cite{pinheiro2014weakly} + sp         & ImNet    & $\sim$760,000& 35.8 & 42.3 \\
\cite{pathak2015constrained} + sz      & VOC aug. &       10,582 & 43.3 & 48.9 \\
\cite{pathak2015constrained} + sz + CRF& VOC aug. &       10,582 & 45.1 & 51.2 \\
\cite{papandreou2015weakly} + CRF      & VOC aug. &       12,031 & 39.6 & 45.2 \\
\hline
\multicolumn{3}{l}{\it Weak supervision} & & \\
\cite{pathak2014fully}                 & VOC aug. &       12,031 & 25.7 & ~~-  \\
\cite{pathak2015constrained}           & VOC aug. &       10,582 & 35.6 & 39.5 \\
\cite{papandreou2015weakly}            & VOC aug. &       12,031 & 35.2 & 40.3 \\
Ours                                   & VOC+YTube&        3,139 & {\bf 39.8} & {\bf 49.6}   \\
Ours                                   & VOC+ImNet&        3,155 & 36.9 & 48.0 \\
\hline
\end{tabular}
\vspace{0.2cm}
\caption{Evaluation on the VOC 2012 test set shown as IoU scores. We compare
with several recent weakly-supervised methods:
EM-Adapt~\cite{papandreou2015weakly}, \cite{pathak2014fully},
\cite{pathak2015constrained}, as well as methods using strong or full
supervision: \cite{pinheiro2014weakly}+bb, \cite{pinheiro2014weakly}+seg,
\cite{papandreou2015weakly}+seg, \cite{mostajabi15,zheng2015conditional}, and
those using additional information: \cite{pinheiro2014weakly}+sp,
\cite{pathak2015constrained}+sz, \cite{pathak2015constrained}+sz+CRF,
\cite{papandreou2015weakly}+CRF.}
\label{tbl:testresults}
\end{center}
\end{table*}
\setlength{\tabcolsep}{1.4pt}

As shown in Table~\ref{tbl:results}, using image data, with additional object
instances, improves the IoU score from 41.2 to 47.2. In comparison, EM-Adapt
re-trained for 10 classes on the original VOC 2012 achieves only 35.8.
Augmenting the dataset with several additional training
images~\cite{HariharanICCV2011}, improves it to 40.2, but this remains
considerably lower than our result. M-CNN trained with ImageNet-VID achieves
39.0 (ImNet in the table), which is comparable to our result with
YouTube-Objects. The performance is significantly lower for the motorbike class
(15.3 vs 32.4) owing to the small number of video shots available for training.
In this case, we only have 67 shots compared to 272 from YouTube-Objects.
Augmenting this dataset with VOC images boosts the performance to 43.7
(VOC+ImNet). Augmenting the training set with additional images (VOC aug.)
further increases the performance.

\paragraph{Qualitative results.} 
\F{fig:usvthem} shows qualitative results of M-CNN (trained on VOC and
YouTube-Objects) on a few sample images. These have much more accurate object
boundaries than the best variant of EM-Adapt~\cite{papandreou2015weakly}, which
tends to localize the object well, but produces a `blob-like' segmentation,
cf.\ last four rows in the figure in particular. The first three rows show
example images containing multiple object categories. M-CNN recognizes object
classes more accurately, e.g., cow in row 5, than EM-Adapt, which confuses cow
(shown in green) with horse (magenta). Furthermore, our segmentation results
compare favorably with the fully-supervised DeepLab~\cite{chen2014semantic}
approach (see rows 4-6), highlighting the impact of motion to learn
segmentation. There is scope for further improvement, e.g., overcoming the
confusion between similar classes in close proximity to each other, as in the
challenging case in row 3 for cat vs dog.

\begin{figure*}
\centering
\fourfigures{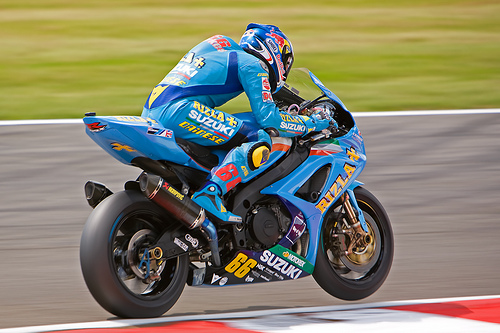}{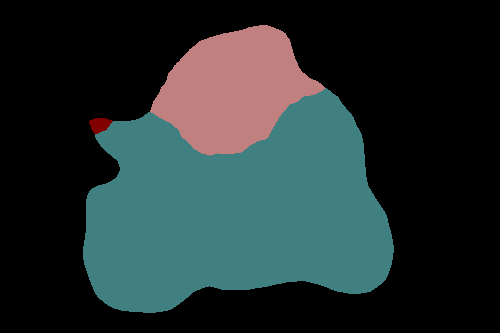}{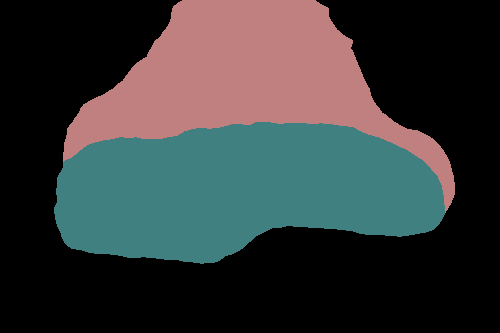}{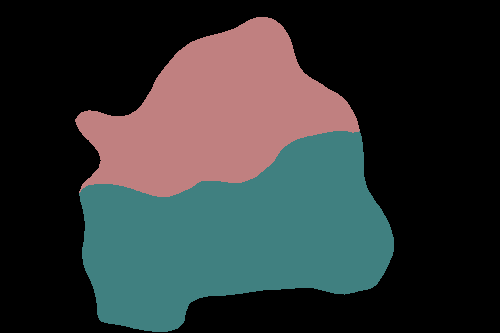}{0.5\columnwidth}
\vspace{0.1cm}
\fourfigures{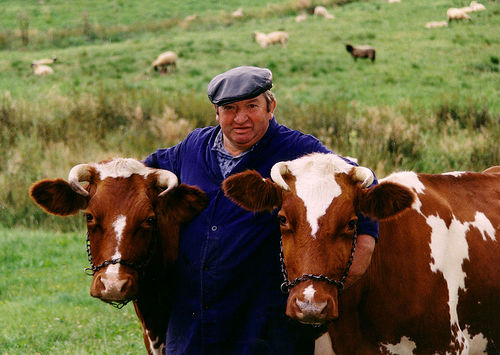}{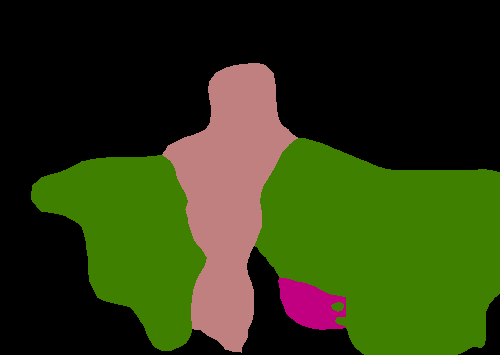}{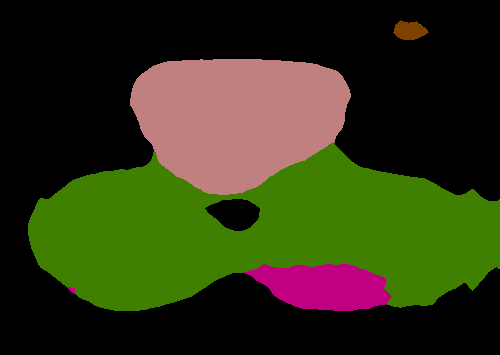}{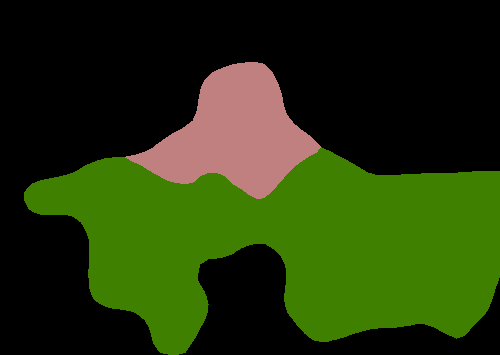}{0.5\columnwidth}
\vspace{0.1cm}
\fourfigures{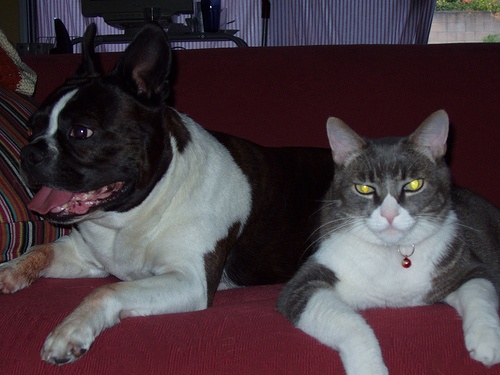}{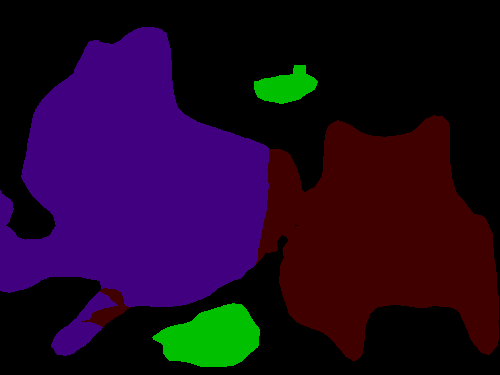}{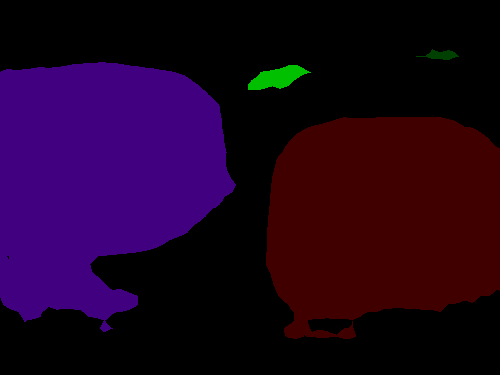}{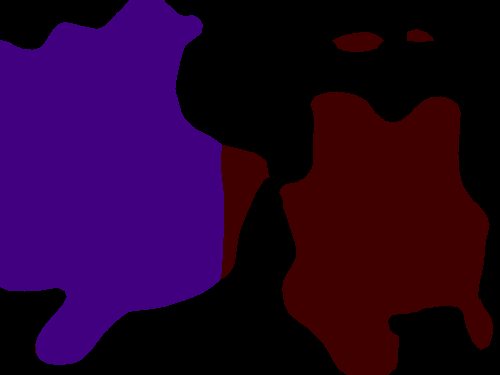}{0.5\columnwidth}
\vspace{0.1cm}
\fourfigures{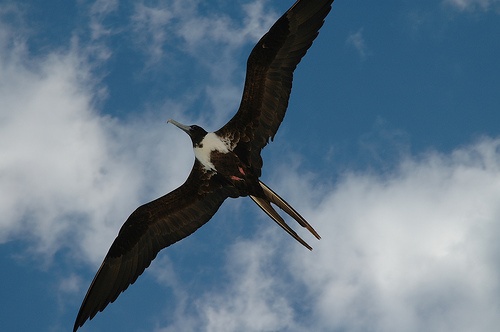}{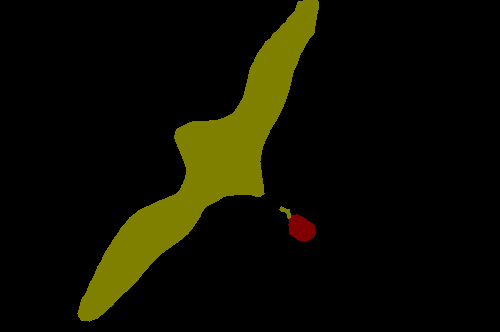}{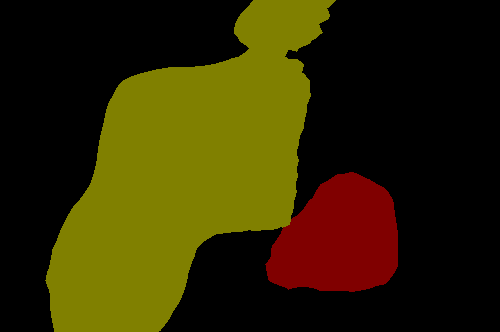}{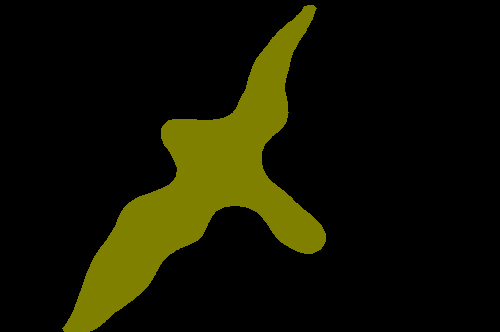}{0.5\columnwidth}
\vspace{0.1cm}
\fourfigures{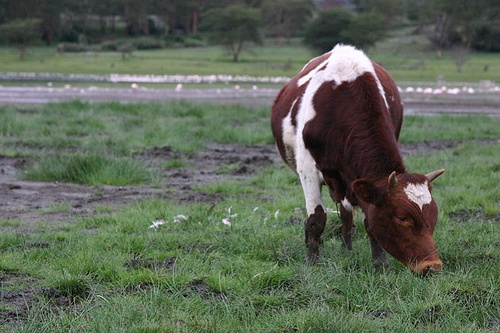}{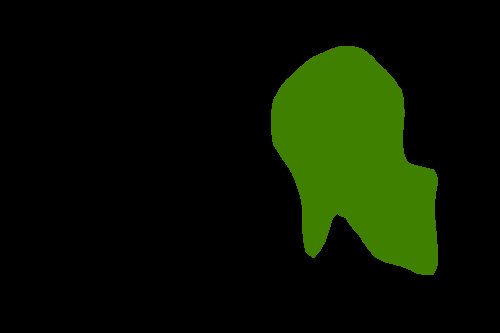}{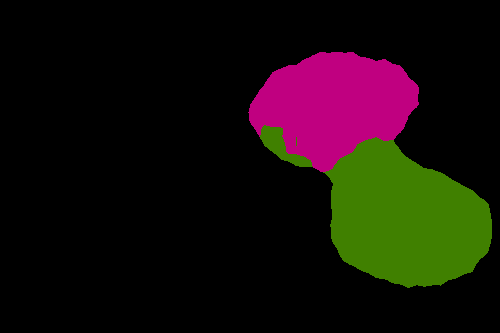}{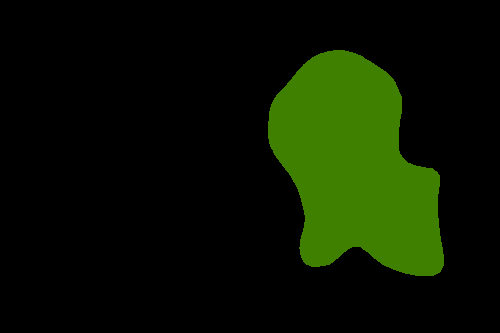}{0.5\columnwidth}
\vspace{0.1cm}
\fourfigures{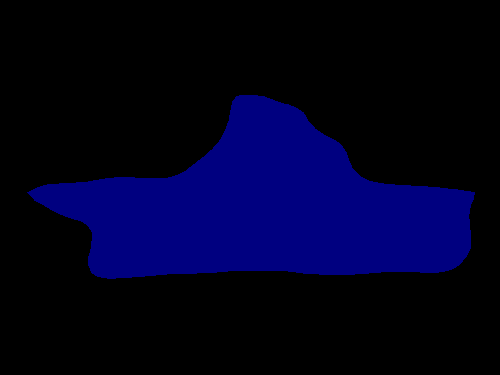}{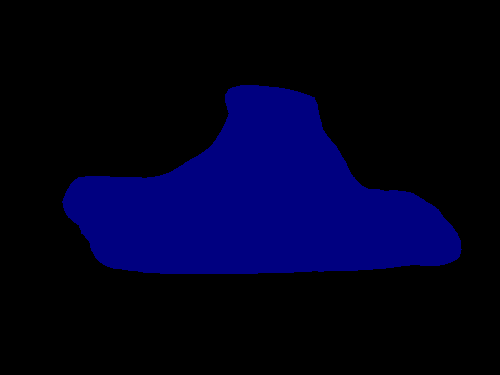}{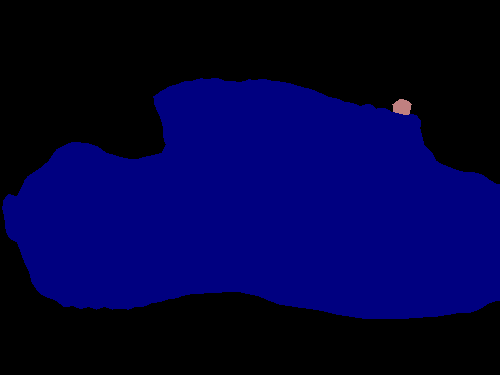}{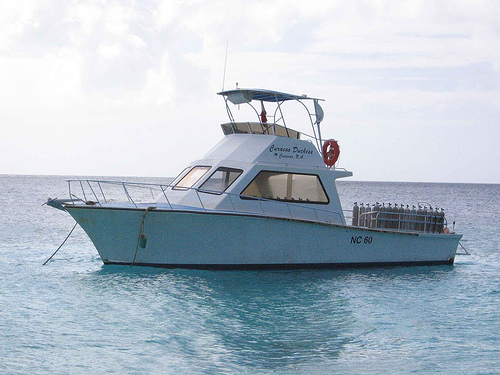}{0.5\columnwidth}
\vspace{0.1cm}
\fourfigures{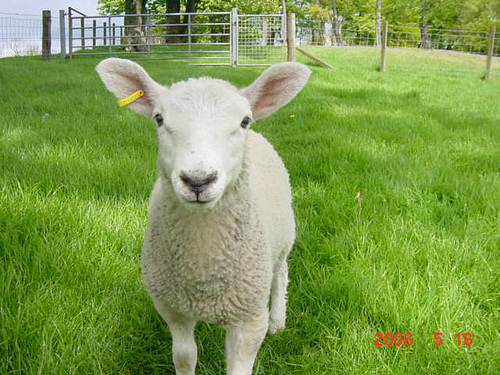}{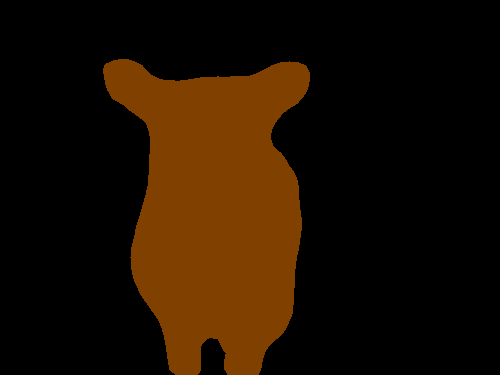}{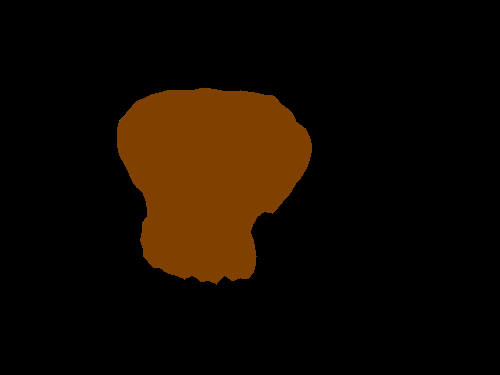}{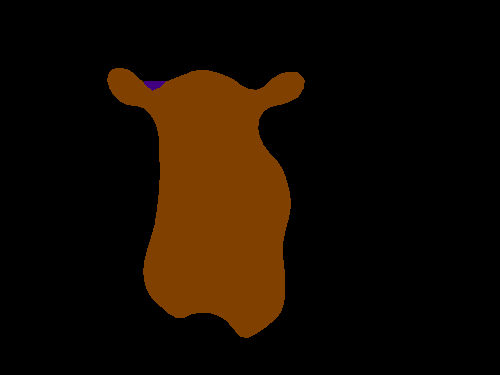}{0.5\columnwidth}
\makebox[0.5\columnwidth][c]{Image~~~}\makebox[0.5\columnwidth][c]{DeepLab~\cite{chen2014semantic} (full)}\makebox[0.5\columnwidth][c]{~~~~~EM-Adapt~\cite{papandreou2015weakly} (weak)}\makebox[0.5\columnwidth][c]{~~~~~~~~M-CNN (weak)}
\vspace{0.2cm}
\caption{Sample results on the VOC 2012 validation set. Results of
fully-supervised DeepLab\cite{chen2014semantic}, weakly-supervised
EM-Adapt~\cite{papandreou2015weakly} trained on augmented VOC, and our
weakly-supervised M-CNN trained on VOC+YouTube-Objects are shown in 2nd, 3rd
and 4th columns respectively. ({\it Best viewed in color.})}
\label{fig:usvthem}
\end{figure*}

\paragraph{Comparison to the state of the art.}
Table~\ref{tbl:testresults} shows our evaluation on the VOC 2012 test set, with
our model trained on 20 classes. We performed this by uploading our
segmentation results to the evaluation server, as ground truth is not publicly
available for the test set. We compare with several state-of-the-art methods
with scores taken directly from the publications,
except~\cite{papandreou2015weakly} without the post-processing CRF step. This
result, shown as `\cite{papandreou2015weakly}' in the table, is with a model we
trained on the VOC augmented dataset. We train M-CNN on all the 20 VOC classes
with the model trained (and fine-tuned) on YouTube-Objects and perform a second
fine-tuning step together with videos from YouTube-Objects and images from VOC.
This achieves 39.8 mean IoU over all the 20 classes, and 49.6 on the 10 classes
with video data. This result is significantly better than recent methods using
only weak labels, which achieve 25.7~\cite{pathak2014fully},
35.6~\cite{pathak2015constrained} and 35.2~\cite{papandreou2015weakly}. The
improvement shown by our M-CNN is more prominent when we consider the average
over 10 classes where we use soft motion segmentation cues (and the
background), with nearly 10\% and 9\% boost over~\cite{pathak2015constrained}
and~\cite{papandreou2015weakly} respectively. We also show the evaluation of
the model trained on ImageNet-VID in the table.

\begin{table*}[t]
\begin{center}
\begin{tabular}{l|C{10mm}C{10mm}C{10mm}C{10mm}C{10mm}C{10mm}C{10mm}C{10mm}C{10mm}C{10mm}|c}
\hline
Method & aero & bird & boat & car & cat & cow & dog & horse & mbike & train & Average \\
\hline
\multicolumn{3}{l}{\it Unsupervised}    &      &      &      &      &      &      &      &      &      \\
\cite{brox2010object}    & 53.9 & 19.6 & 38.2 & 37.8 & 32.2 & 21.8 & 27.0 & 34.7 & 45.4 & 37.5 & 34.8 \\
\cite{papazoglou2013fast}& 65.4 & 67.3 & 38.9 & 65.2 & 46.3 & 40.2 & 65.3 & 48.4 & 39.0 & 25.0 & 50.1 \\
\cite{Kwak15}            & 55.2 & 58.7 & 53.6 & 72.3 & 33.1 & 58.3 & 52.5 & 50.8 & 45.0 & 19.8 & 49.9 \\
\hline
\multicolumn{4}{l}{\it Weakly supervised}      &      &      &      &      &      &      &      &      \\
\cite{prest2012learning} & 51.7 & 17.5 & 34.4 & 34.7 & 22.3 & 17.9 & 13.5 & 26.7 & 41.2 & 25.0 & 28.5 \\
\cite{Joulin14}          & 25.1 & 31.2 & 27.8 & 38.5 & 41.2 & 28.4 & 33.9 & 35.6 & 23.1 & 25.0 & 31.0 \\
\cite{Kwak15}            & 56.5 & 66.4 & 58.0 & 76.8 & 39.9 & 69.3 & 50.4 & 56.3 & 53.0 & 31.0 & 55.7 \\
M-CNN                    & 76.1 & 57.7 & 77.7 & 68.8 & 71.6 & 75.6 & 87.9 & 71.9 & 80.0 & 52.6 & {\bf 72.0} \\
\hline
\end{tabular}
\vspace{0.2cm}
\caption{Co-localization performance of M-CNN. We report per class CorLoc
scores, and compare with state-of-the-art
unsupervised~\cite{papazoglou2013fast,brox2010object,Kwak15} and weakly
supervised~\cite{Kwak15,prest2012learning,Joulin14} methods. See text for
details.}
\label{tbl:corloc}
\end{center}
\end{table*}

A few methods have used additional information in the training process, such as
the size of objects (+ sz in the table), superpixel segmentation (+ sp), or
post-processing steps, e.g., introducing a CRF with pairwise terms learned from
fully-annotated data (+ CRF), or even strong or full supervision, such as
bounding box (+ bb) or pixel-level segmentation (+ seg) annotations. Even
though our pure weakly-supervised method is not directly comparable to these
approaches, we have included these results in the table for completeness.
Nevertheless, M-CNN outperforms some of these
methods~\cite{pinheiro2014weakly,papandreou2015weakly}, due to our effective
learning scheme. Also from Table~\ref{tbl:testresults}, the number of training
samples used for M-CNN (number of videos shots + number of VOC training images)
is significantly lower than those for all the other methods.

\subsection{Co-localization}
\label{sec:corloc}
We perform co-localization in the standard setting, where videos contain a
common object. Here, we use our M-CNN trained on the YouTube-Objects dataset
with 10 categories. We evaluate it on all the frames in YouTube-Objects to
obtain prediction scores ${\bf p}_i$ for each pixel $i$. With these scores, we
compute a foreground GMM by considering pixels with high predictions for the
object category as foreground. A background GMM is also computed in a similar
fashion. These form the unary term $\psi^m_i$ in the energy function
(\ref{eqn:energy}). We then minimize this function with graph cut based
inference to compute the binary (object vs background) segmentation labels.
Since we estimate segmentations for all the video frames, we do this at the
superpixel level~\cite{Achanta12} to reduce computation cost. We then extract
the bounding box enclosing the largest connected component in each frame, and
evaluate them following~\cite{prest2012learning}. Quantitative results with
this are summarized as per-class and average CorLoc scores in
Table~\ref{tbl:corloc}. We observe that our result outperforms previous state
of the art~\cite{Kwak15} by over 16\%. Performing this experiment with
ImageNet-VID data we obtain 42.1 on average, in comparison to 37.9
of~\cite{papazoglou2013fast}. ImageNet-VID being a more challenging dataset
than YouTube-Objects results in a lower performance for both these methods.

\begin{figure*}
\begin{center}
\threefigures{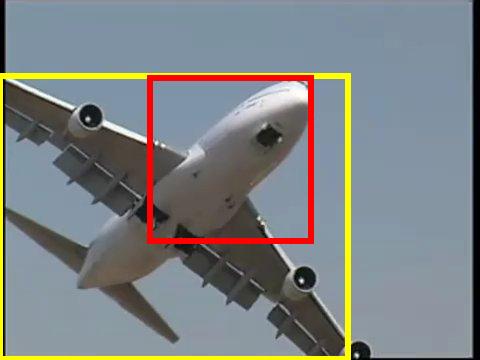}{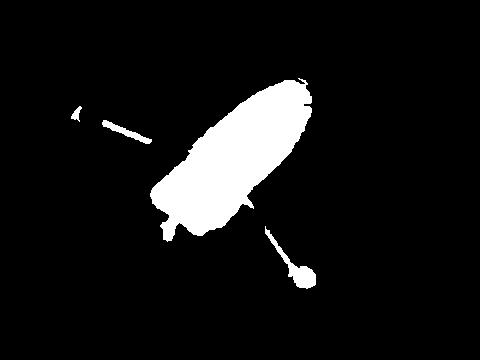}{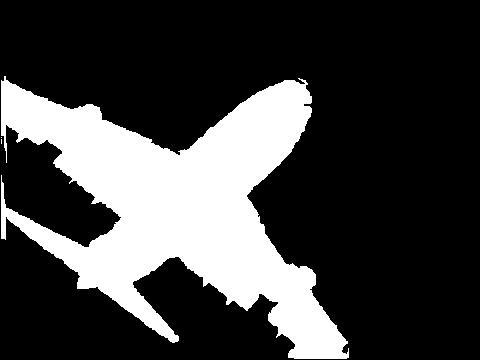}{0.6\columnwidth} 
\vspace{0.1cm}
\threefigures{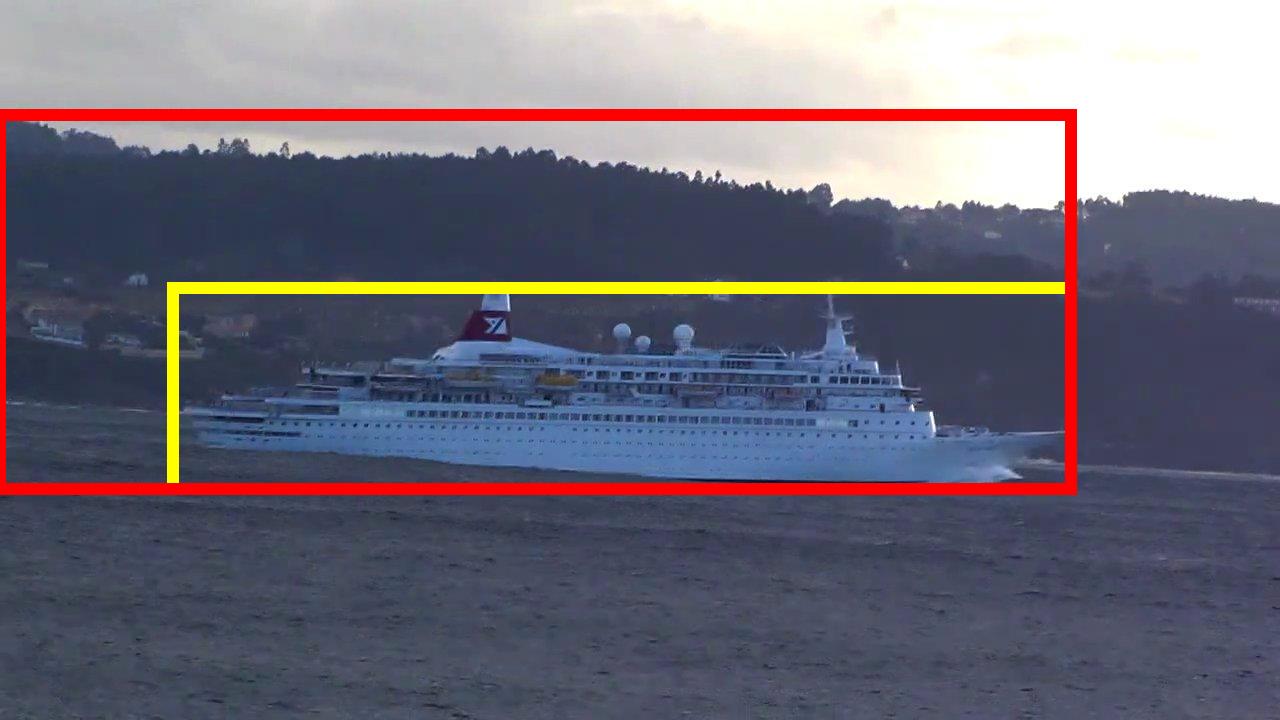}{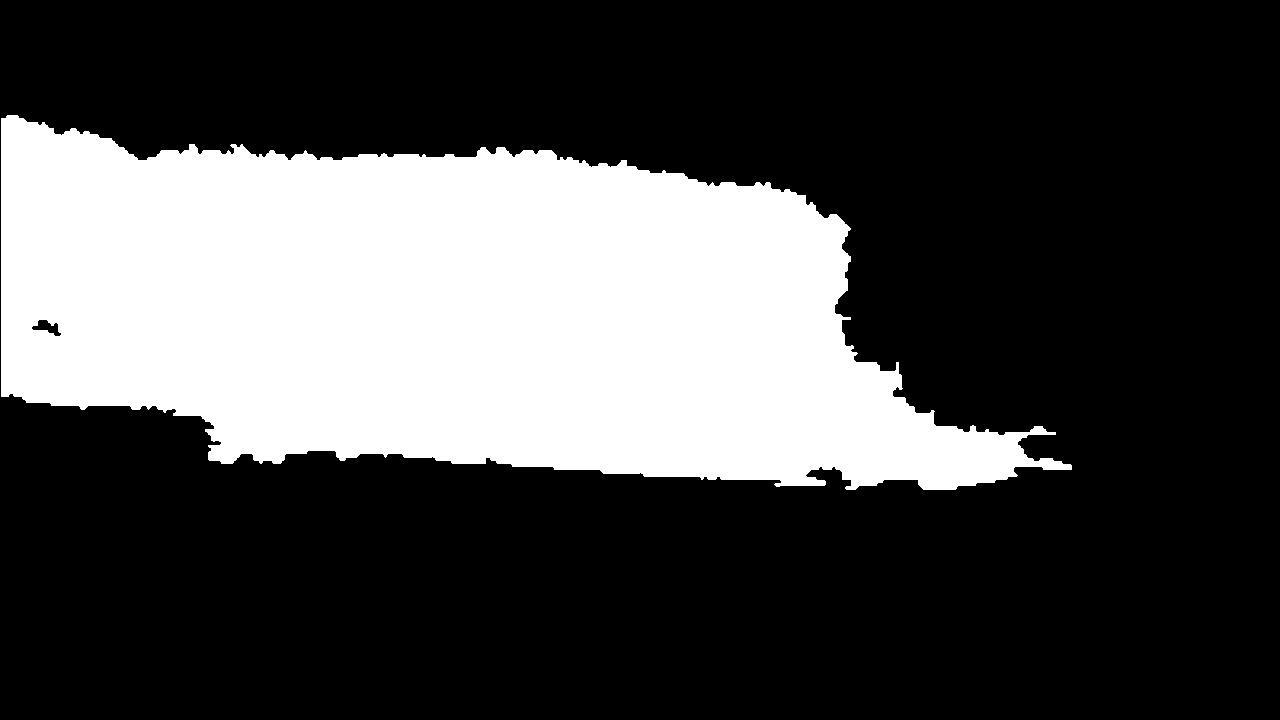}{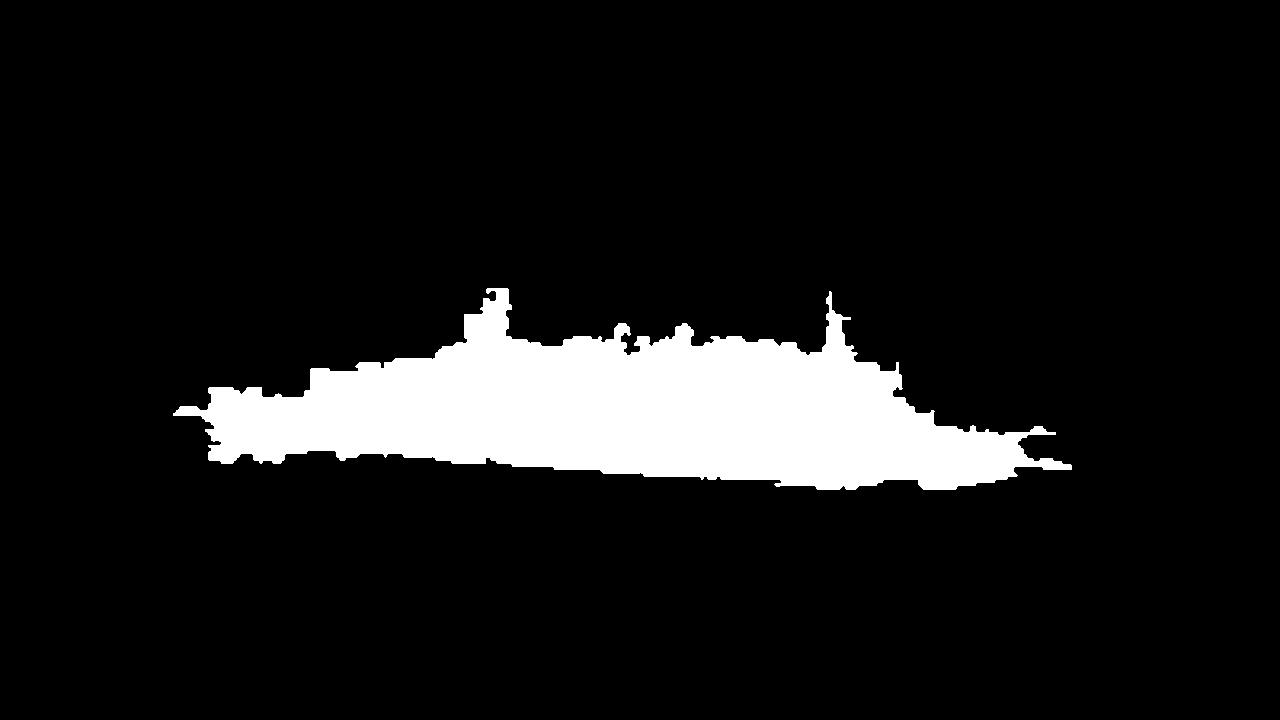}{0.6\columnwidth} 
\vspace{0.1cm}
\threefigures{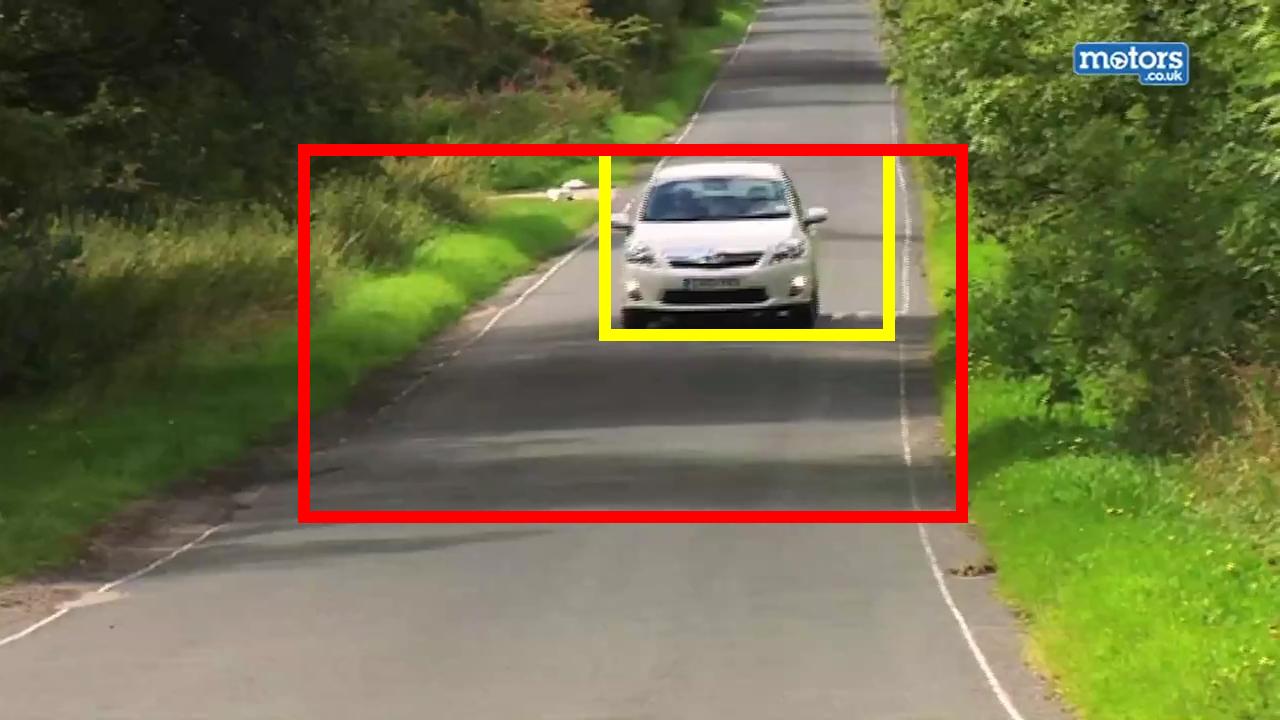}{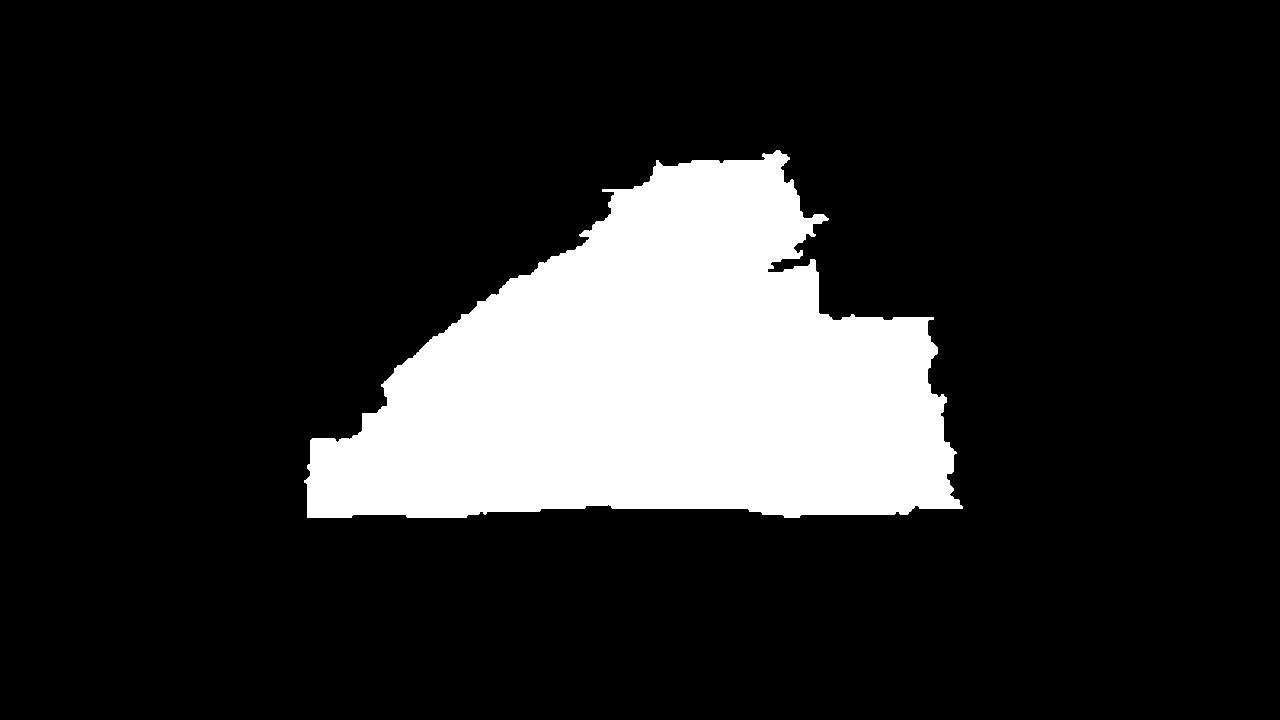}{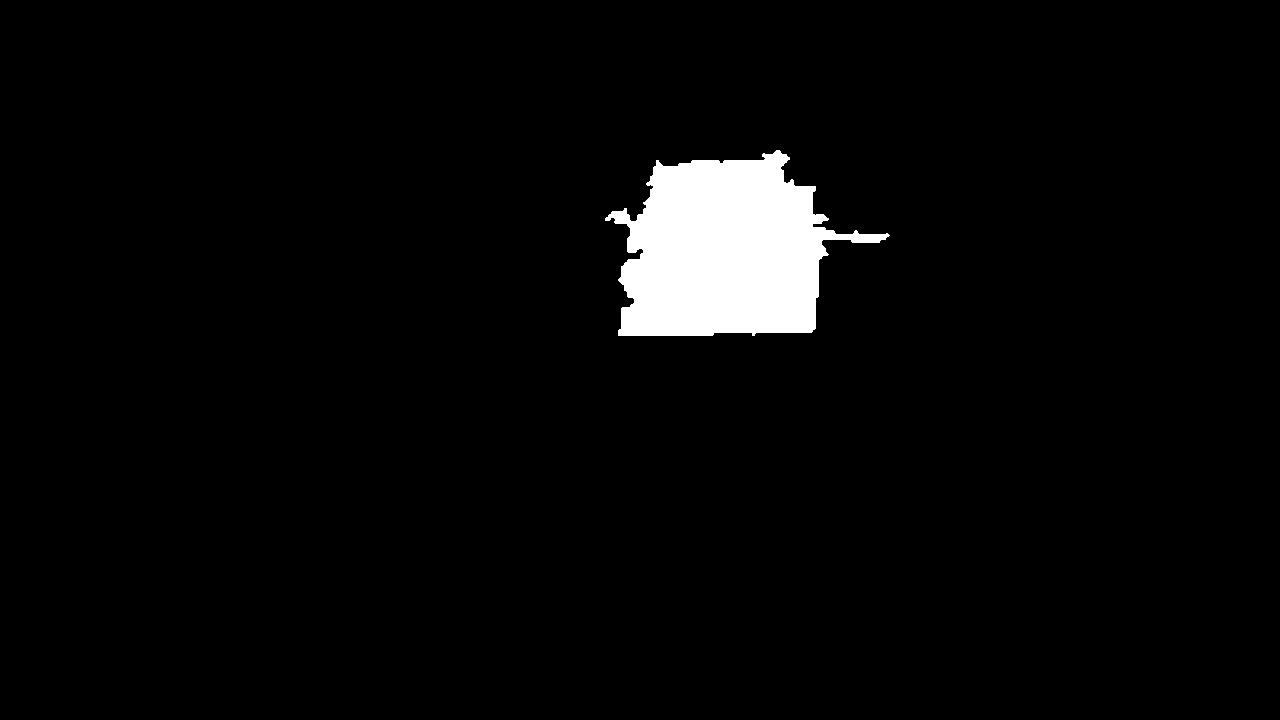}{0.6\columnwidth} 
\vspace{0.1cm}
\threefigures{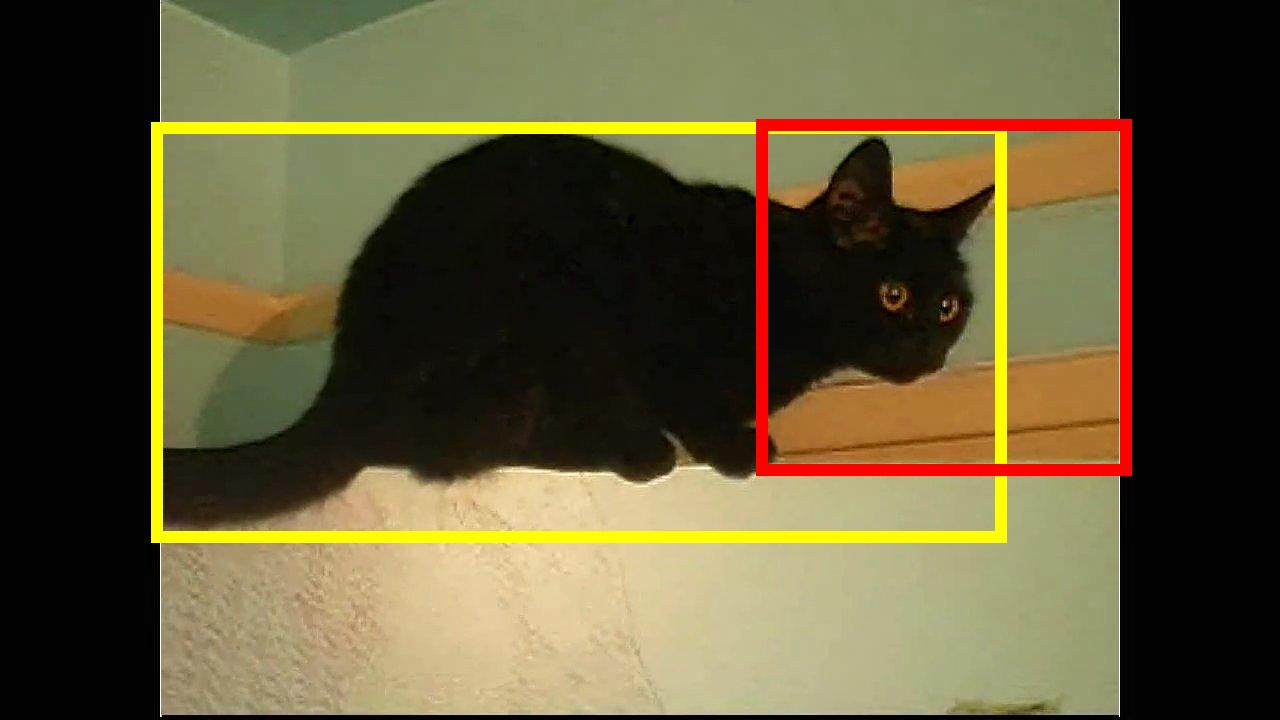}{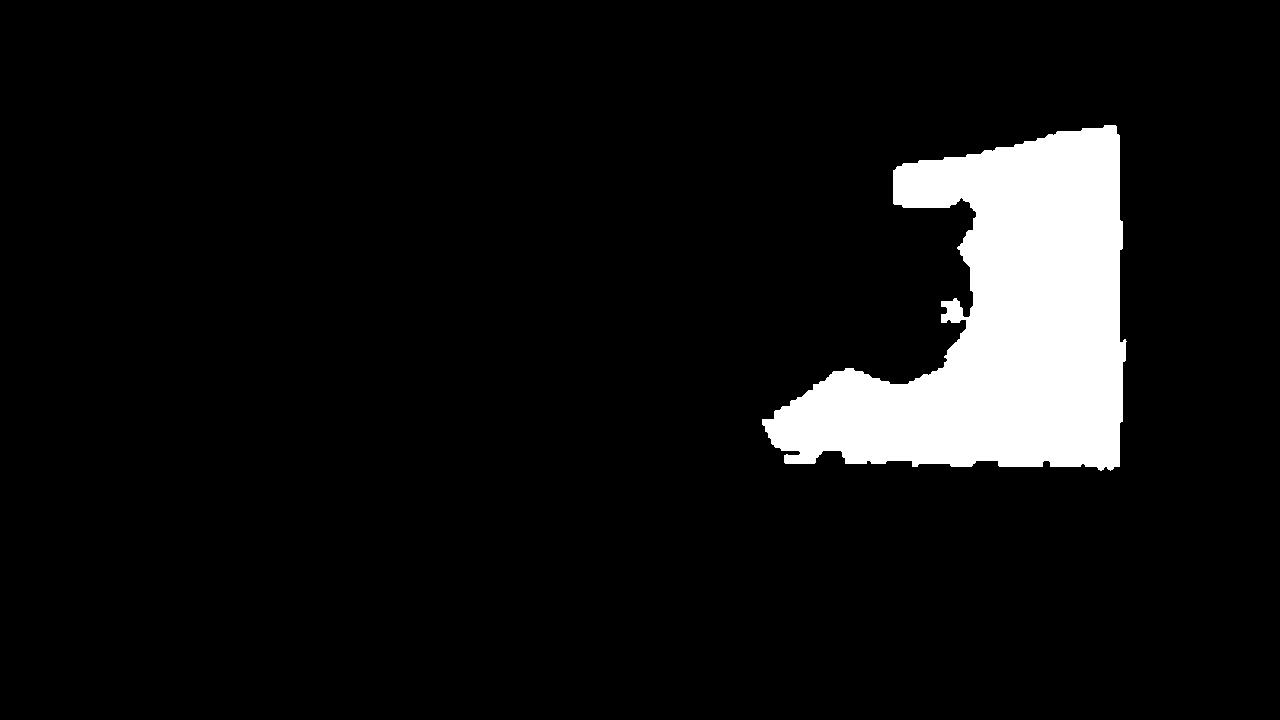}{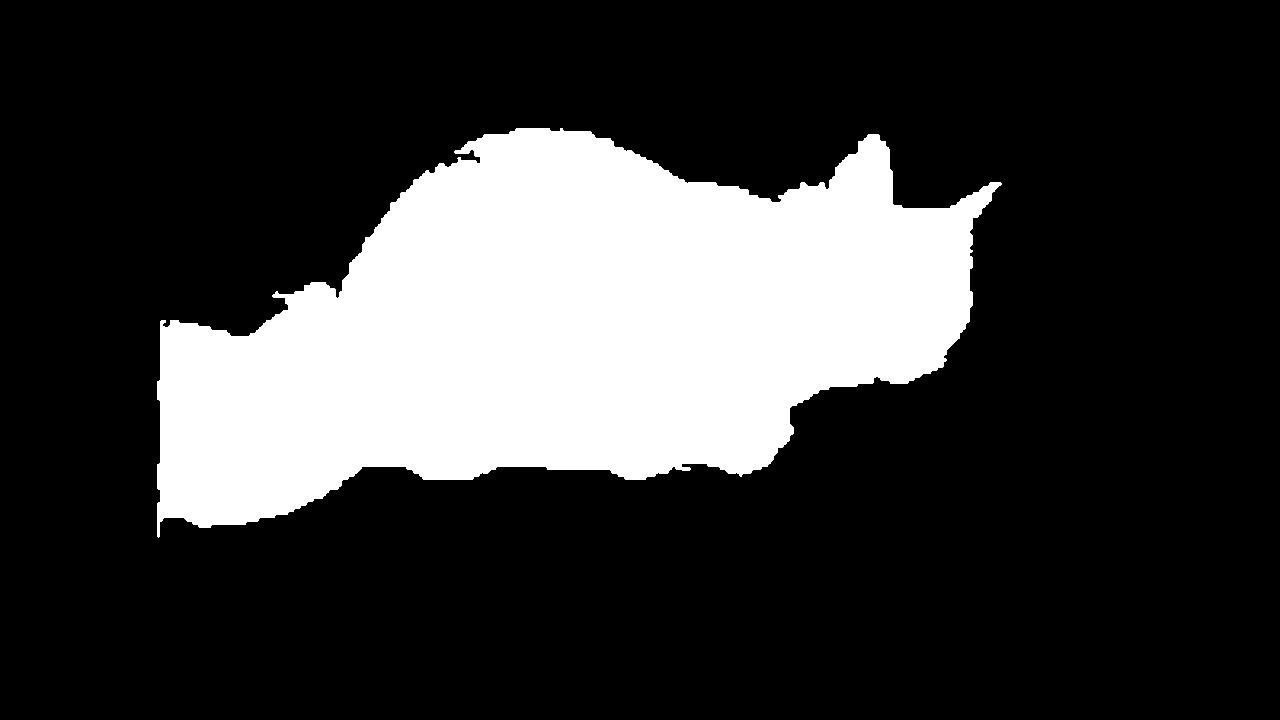}{0.6\columnwidth} 
\vspace{0.1cm}
\threefigures{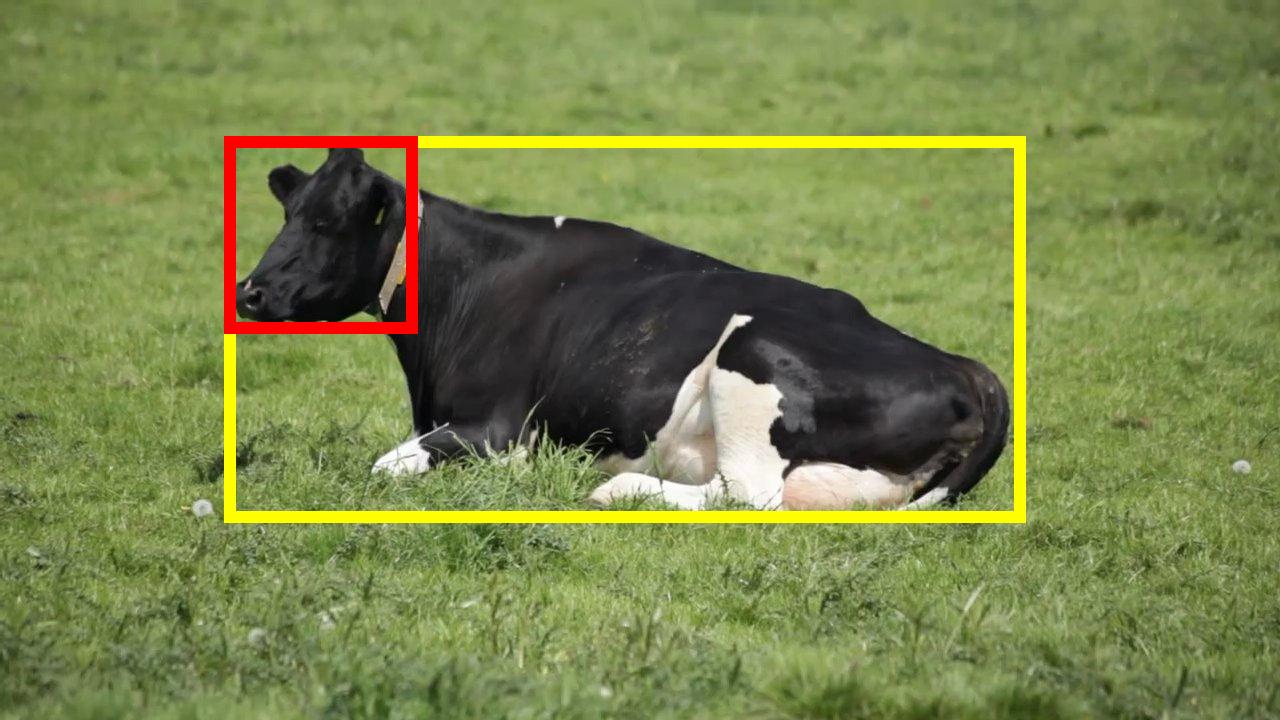}{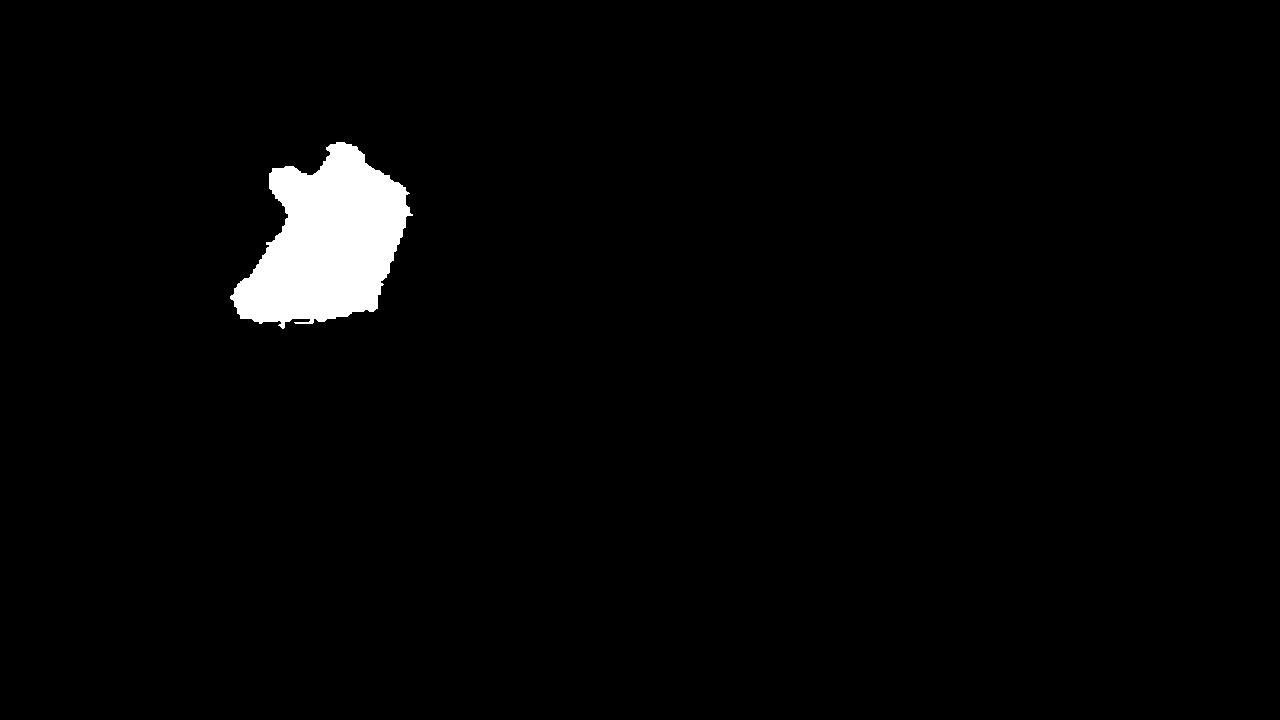}{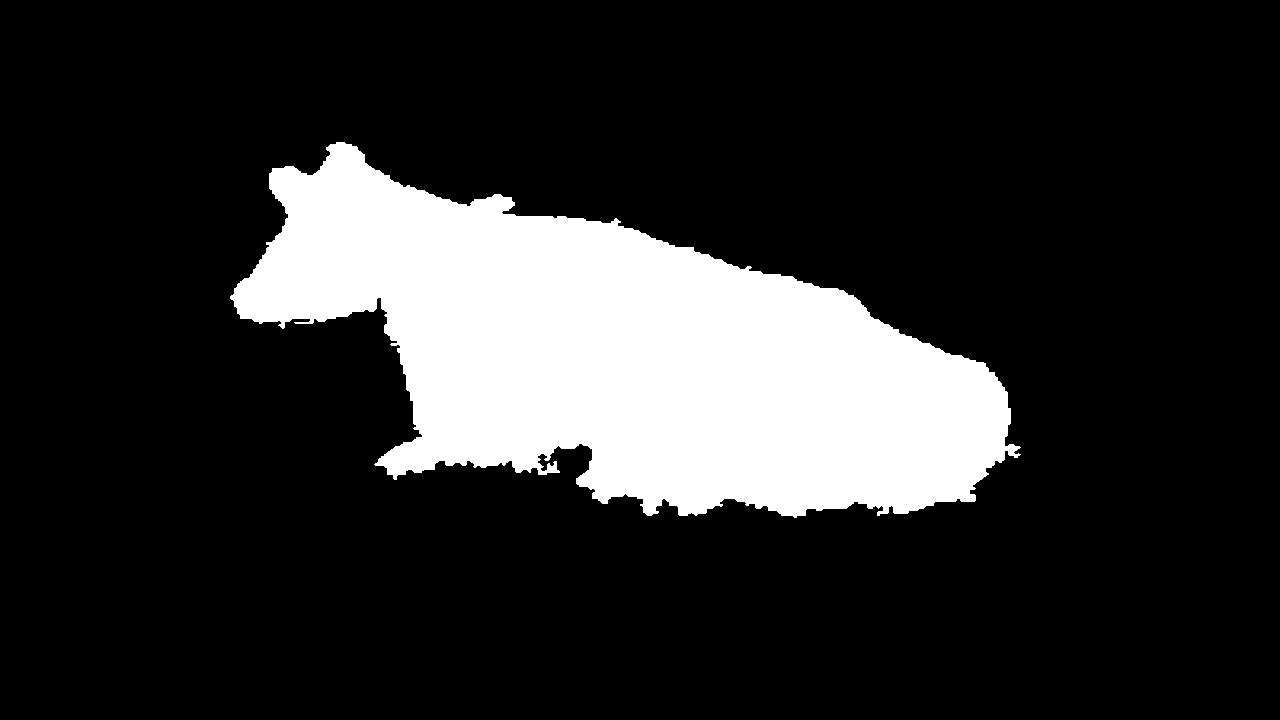}{0.6\columnwidth} 
\vspace{0.1cm}
\threefigures{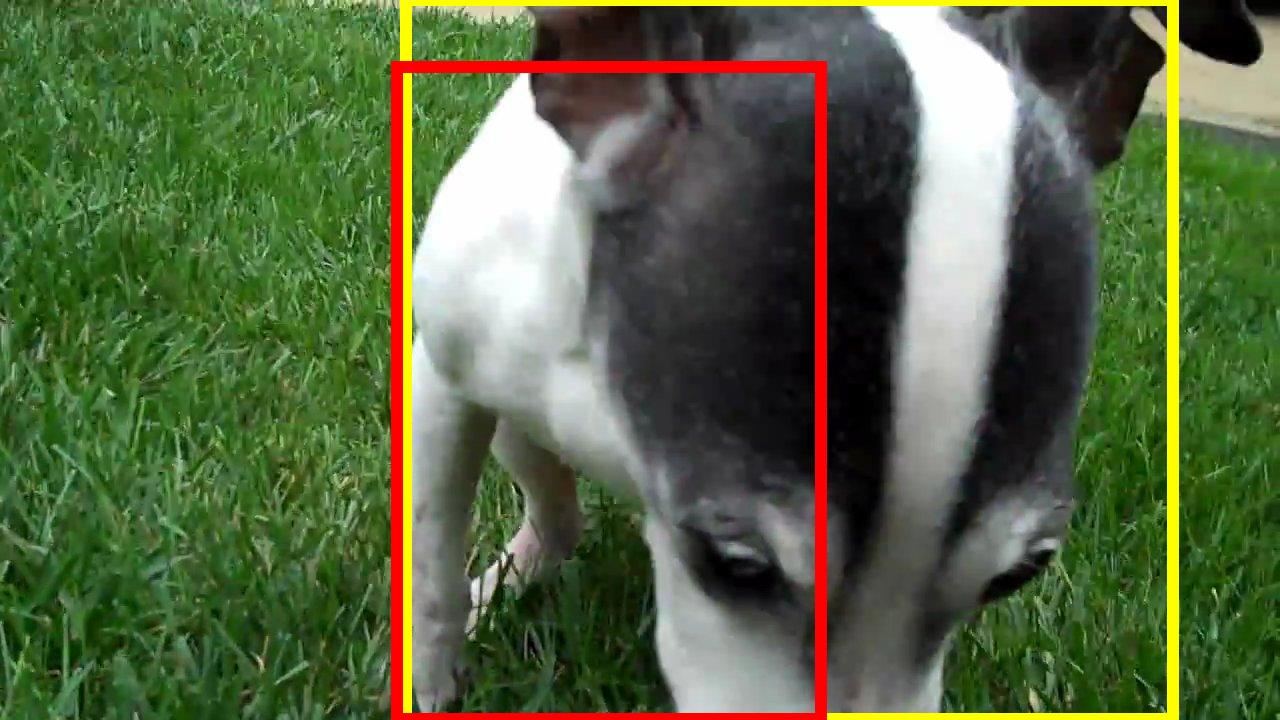}{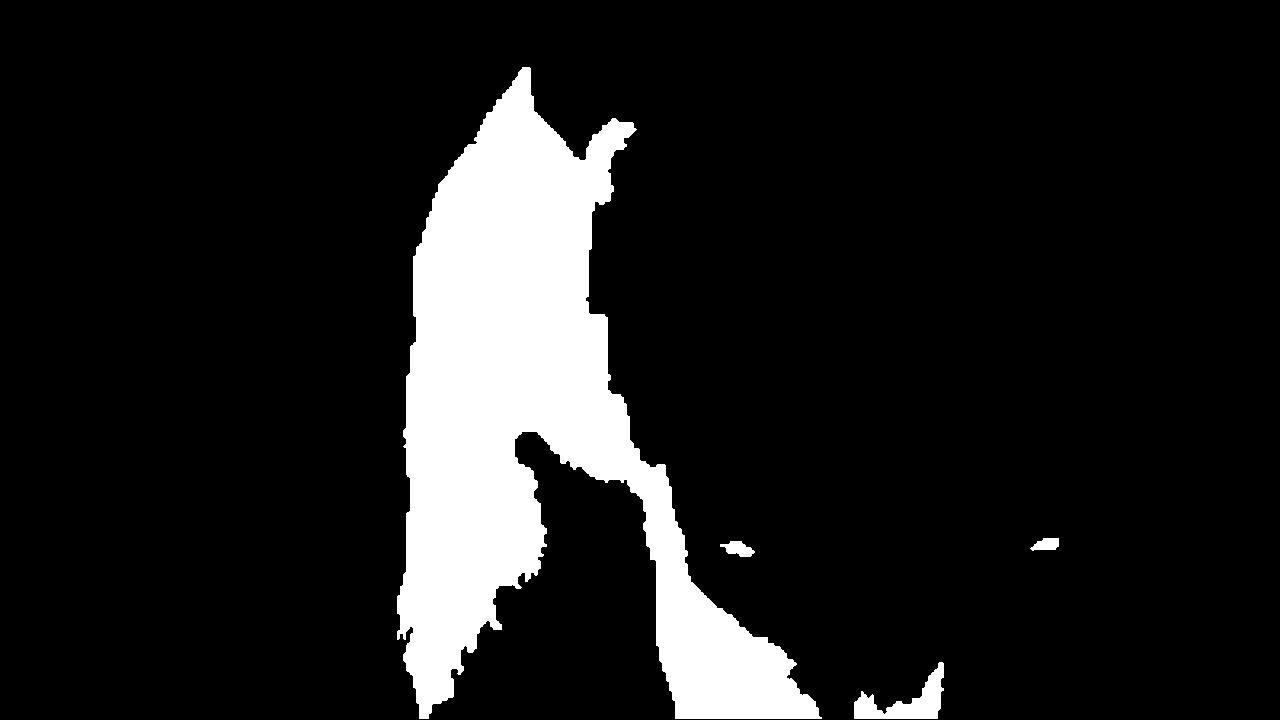}{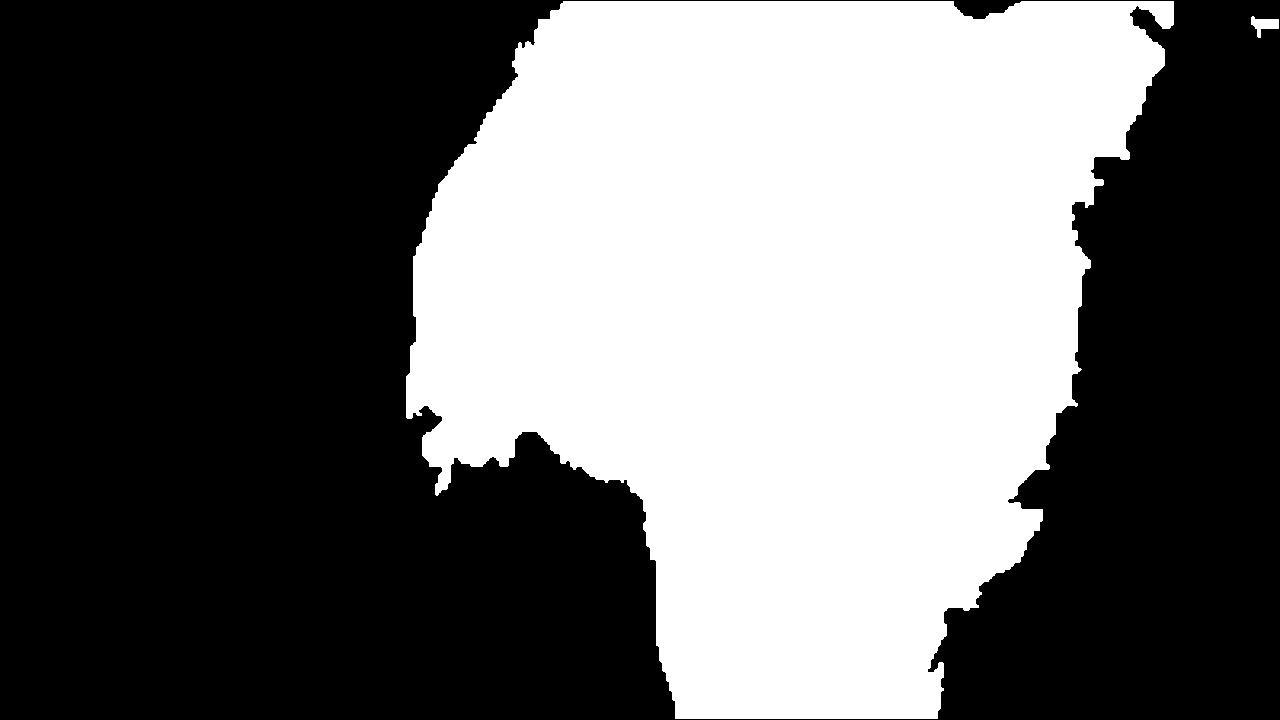}{0.6\columnwidth} 
\vspace{0.1cm}
\threefigures{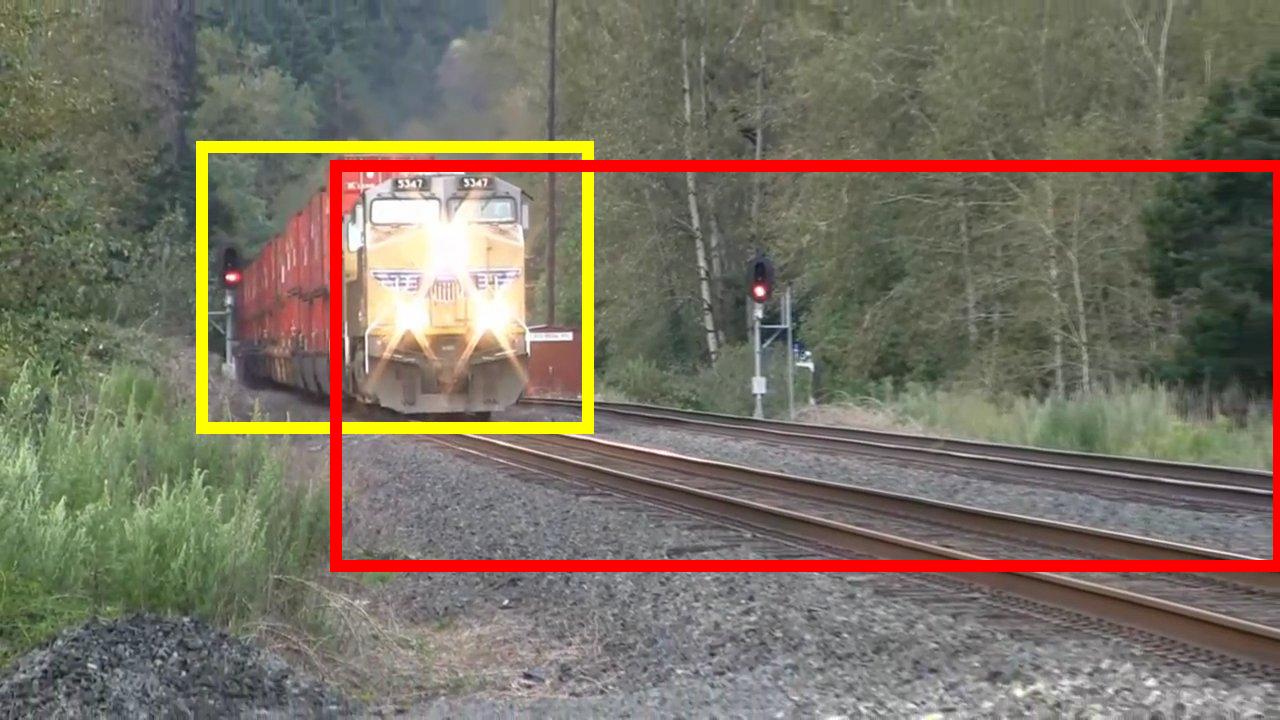}{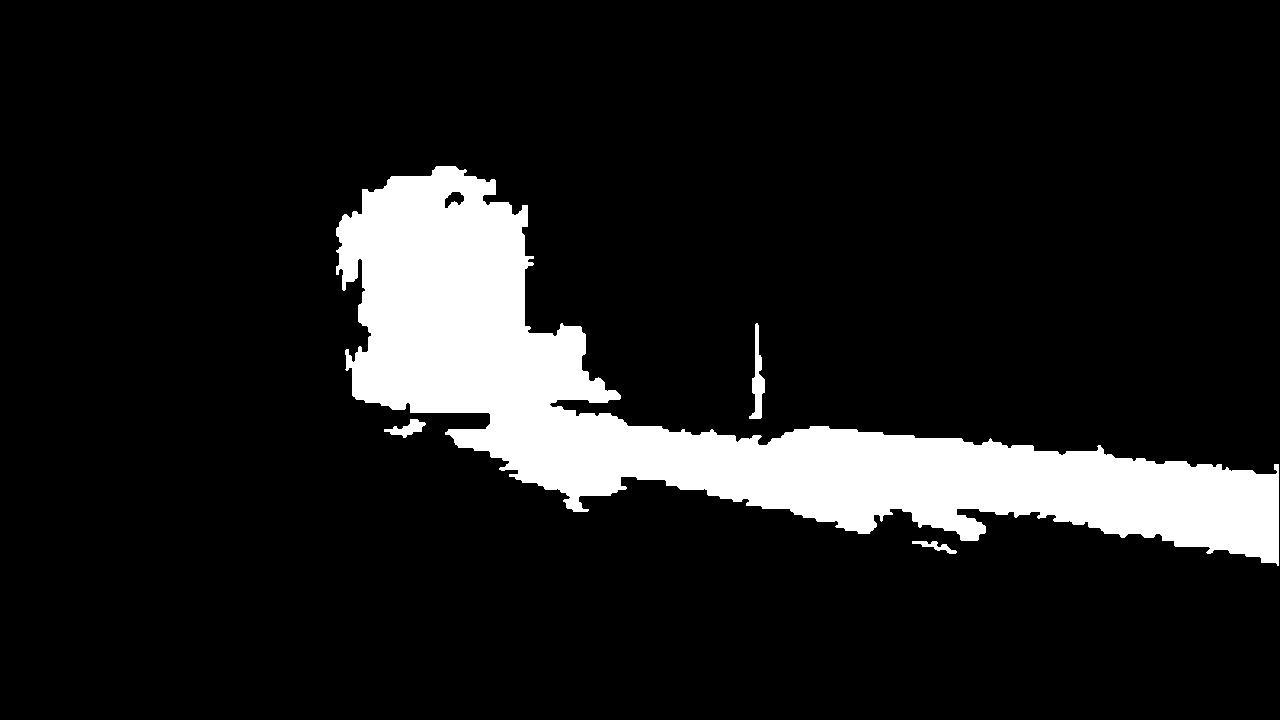}{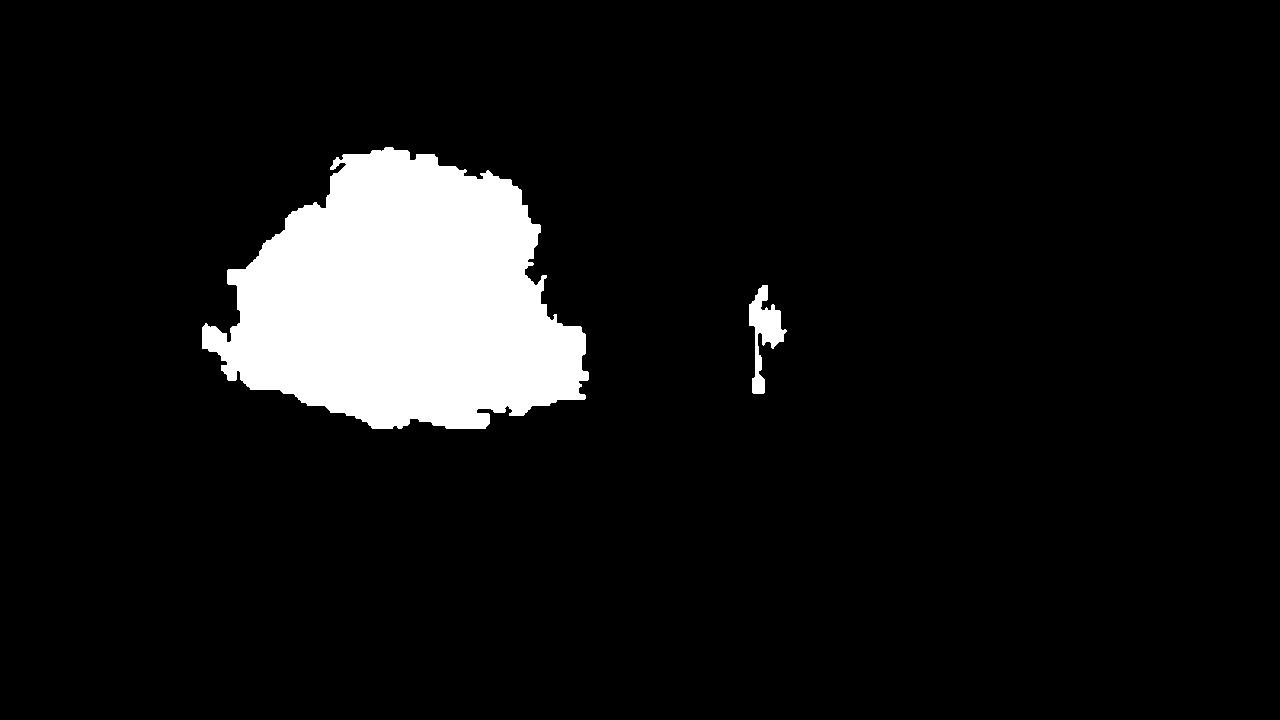}{0.6\columnwidth} 
\makebox[0.6\columnwidth][c]{Frame~~~}\makebox[0.6\columnwidth][c]{~Motion seg.~\cite{papazoglou2013fast}}\makebox[0.6\columnwidth][c]{~~~~~M-CNN}
\end{center}
\caption{Sample co-localization results on the YouTube-Objects dataset. In the
first column the estimated bounding boxes are shown, where yellow corresponds
to our result, and red to that of~\cite{papazoglou2013fast}. Segmentations
corresponding to~\cite{papazoglou2013fast} and our method are shown in columns
2 and 3 respectively. ({\it Best viewed in color.})}
\label{fig:cor_loc}
\end{figure*}

We qualitatively demonstrate the performance of our method on the
YouTube-Objects dataset in Figure~\ref{fig:cor_loc}. Our method produces stable
results on a variety of categories (third column in the figure). The
performance of the motion segmentation method~\cite{papazoglou2013fast} is also
shown for comparison. It is limited by the quality of optical flow and the
heuristics used to distinguish foreground from background motion. As a result,
it often fails, see second column in the figure.

\section{Summary}
\label{sec:summary}
This paper introduces a novel weakly-supervised learning approach for semantic
segmentation, which uses only class labels assigned to videos. It integrates
motion cues computed from video as soft constraints into a fully convolutional
neural network. Experimental results show that our soft motion constraints can
handle noisy motion information and improve significantly over the heuristic
size constraints used by state-of-the-art approaches for weakly-supervised
semantic segmentation, i.e., by EM-Adapt~\cite{papandreou2015weakly}. We show
that our approach outperforms previous state of the
art~\cite{pathak2015constrained,papandreou2015weakly} on the PASCAL VOC 2012
image segmentation dataset, thereby overcoming domain-shift issues typically
seen when training on video and testing on images.  Furthermore, our
weakly-supervised method shows excellent results for video co-localization and
improves significantly over several recent
methods~\cite{Kwak15,Joulin14,papazoglou2013fast}.

\paragraph{Acknowledgments.} This work was supported in part by the ERC
advanced grant ALLEGRO, the MSR-Inria joint project, a Google research award
and a Facebook gift. We gratefully acknowledge the support of NVIDIA with the
donation of GPUs used for this research.

\bibliographystyle{ieee}
\bibliography{mcnn}

\end{document}